\documentclass[12pt,a4paper]{article}

\usepackage{graphics, graphicx, color, latexsym, epsfig, verbatim, amssymb,  lscape, amsmath, amsbsy, amstext, natbib, turnstile, amsfonts, dsfont, mathrsfs, multirow}

\usepackage[ruled]{algorithm2e}                 

\newtheorem{thm}{Theorem}[section]

\newtheorem{defi}{Definition}[section]
\newtheorem{cor}{Corollary}[section]
\newtheorem{ass}{Assumption}[section]

\newcommand{\be}{\begin{equation}}
\newcommand{\ee}{\end{equation}}
\newcommand\bes{\begin{eqnarray}}
\newcommand\ees{\end{eqnarray}}
\newcommand\besn{\begin{eqnarray*}}
\newcommand\eesn{\end{eqnarray*}}

\newcommand \vc[1]{{\mbox{\boldmath${#1}$}}}

\newcommand \vtheta{\vc \theta}

\newcommand \vZ{\vc z}

\numberwithin{equation}{section}

\newcommand{\norm}[1]{\left\Vert#1\right\Vert}  

\newcommand{\sbr}[1]{\left(#1\right)}        
\newcommand{\mbr}[1]{\left[#1\right]}        
\newcommand{\bbr}[1]{\left\{#1\right\}}      

\def\eop{{\hfill\vbox{\hrule height .3pt
      \hbox{\vrule width.3pt height 7pt
      \kern 7pt
      \vrule width .3pt}
      \hrule height .3pt}} \par\bigskip}

\usepackage{numcompress}

\title{{Communication-efficient Byzantine-robust distributed learning with statistical guarantee}}
\author{Xingcai Zhou, Le Chang, Pengfei Xu and Shaogao Lv\\
{\sl School of Statistics and Mathematics}\\
{\sl Nanjing Audit University, Nanjing, 211085, Jiangsu, China}}
\date{}

\begin{document}
\maketitle
\setcounter{page}{1}

\begin{abstract}
Communication efficiency and robustness are two major issues in modern distributed learning framework. This is due to the practical situations where some computing nodes may have limited communication power or may behave adversarial behaviors. To address the two issues simultaneously,
this paper develops two communication-efficient and robust distributed learning algorithms for convex problems. Our motivation is based on surrogate likelihood framework and the median and trimmed mean operations. Particularly, the proposed algorithms are provably robust against Byzantine failures, and also achieve optimal statistical rates for strong convex losses and convex (non-smooth) penalties. For typical statistical models such as generalized linear models, our results show that statistical errors dominate optimization errors in finite iterations.
Simulated and real data experiments are conducted to demonstrate the numerical performance of our algorithms.
\end{abstract}


\section{Introduction}

In many real-world applications, such as computer vision, natural language processing and recommendation systems, the exceedingly large size of data has made it impossible to store all of them on a single machine. Now, more and more data are stored locally in  individual agents' or users' devices. Statistical analysis in modern era has to deal with the distributed storage data, which
faces tremendous challenge on statistical method, computation and communication.

In several practical situations,  smart-phone or remote devices with limited communication powers may sever as local nodes, or sometimes communication decay  occurs from the constraint of network bandwidth  \citep{KoneMY2016}.
To address  communication issue, communication efficiency-oriented algorithms for distributed optimization have been the focus of amounts of works in the past several years, for example,
\cite{ZhangDuchiW2013}, \cite{Shamiretal2014}, \cite{WangKolarSZ2017}, \cite{JordanLeeYang2019} and among others. This literature has focused on data-parallel mode in which the overall dataset is partitioned and stored on $m$ worker machines that are processed independently.
Among those existing distributed approaches, the divide and conquer may be the simplest strategy with a single communication round, where a master machine takes responsible to ultimately aggregate all local results computed independently at each local worker.

Although the divide and conquer strategy has been proved to achieve optimal estimation rates for parametric models and kernel methods \citep*{ZhangDuchiW2013,ZhangDuchiW2015}, the global estimator based on the naive average may not inherent some useful structures from the model, such as sparsity. Moreover,   a lower bound of the sample size $(m)$ assigned at the local nodes is required to attain the optimal statistical rates, that is, $m=\Omega(\sqrt{N})$, where $N$ is the total sample size.  This  deviates from some practical scenarios where the dataset with the size $\sqrt{N}$ is also too large to store at a single node/machine.   In addition,  existing numerical analysis \citep*{JordanLeeYang2019} have shown that  the naive averaging often performs poorly for nonlinear models, and even its generalization performance  is usually unreliable when
the local sample sizes among workers differ significantly \citep{FanGuoWang2019}.


In the distributed learning literature for communication efficiency, most of existing works on distributed machine learning consist of  two categories: 1) how to design communication efficient algorithms to reduce the round of communications among
workers \citep*{JordanLeeYang2019,KoneMY2016,LeeLinMY2017,Shamiretal2014}; 2) how to choose a suitable (lossy) compression for broadcasting  parameters \citep*{WangWangS2017}.
Notably, \cite{JordanLeeYang2019} and \cite{WangKolarSZ2017} independently propose a Communication-efficient Surrogate Likelihood (CSL) framework for solving  regular M-estimation problems, which also works for high-dimensional penalized regression and Bayesian statistics. Under the master-worker architectures, CSL makes full use of the total information of the data over the master  machine, while only merges the first-order gradients from all the workers.
Specially, a quasi-newton optimization at the master is solved as the final estimator, instead of merely aggregating all the local estimators like one-shot methods. It has been shown in \citep*{JordanLeeYang2019,WangKolarSZ2017} that
  CSL-based distributed learning can preserve sparsity structure and achieve optimal statistical estimation rates for convex problems in finite-step iterations.

Despite the generality and elegance of the CSL framework, it is not a wisdom that if it would be directly applied  to   Byzantine learning. In view that CSL aggregation rule heavily depends on the local gradients, the learning performance will be degraded significantly  if these received gradients from local workers are highly noisy.
In fact, Byzantine-failure is frequently encountered in distributed or federated learning \citep{YinChenRB2018}. In a decentralized environment, some computing units may exhibit abnormal behavior due to crashes, stalled computation or unreliable communication channels. It is typically modeled as Byzantine failure, meaning that some worker machines may behave arbitrary and potentially adversarial behavior. Thus, it leads to the misleading learning process \citep{VempatyTongV2013,YangGangB2019,WuLingChenG2019}. Robustifying learning against Byzantine failures has attracted a great of attention in recent years.



To copy with Byzantine failures in distributed statistical learning, most resilient approaches in a few recent works tend to combine stochastic gradient descent (SGD) with different robust aggregation rules, such as geometric median \citep{Minsker2015, ChenSuXu2017}, median \citep{XieKoyejoG2018, YinChenRB2018}, trimmed mean \citep{YinChenRB2018}, iterative filtering \citep{SuXu2018} and Krum \citep{BlanchardMhamdiGS2017}. These learning algorithms can tolerate a small number of devices attacked by Byzantine adversaries. \cite{YinChenRB2018} developed two distributed learning algorithms that were provably robust against the Byzantine failures, and also these proposed algorithms can achieve optimal statistical error rates for strongly convex losses. Yet, the above works did not consider the communication cost issue and an inappropriate robust technique can result in increasing the number of communications.


In this paper, we  develop two efficient distributed learning algorithms with both communication-efficiency and Byzantine-robustness,  in pursuit of
accurate statistical estimators. The proposed algorithms integral the framework of CSL for effective communication with two robust techniques, which will be described in Section 3. At each round,  the 1st non-Byzantine machine needs to solve a regularized M-estimation problem on its local data. Other workers only need to compute the gradients on their individual data, and then send these local gradients to the 1st non-Byzantine machine. Once receiving these gradient values from the workers,  the 1st non-Byzantine machine further aggregates them  on basis of coordinate-wise median or coordinate-wise trimmed mean technique,  so as to formulate a robust proxy of the  global gradient.

In our communication-efficient and Byzantine-robust framework, our estimation error indicates that there exist several trade offs between statistical efficiency, computation efficiency and robustness. In particular, our algorithms attempt to guard against Byzantine failures meanwhile  not sacrifice the quality of learning. Theoretically, we show the first algorithm achieves the following statistical error rates
$$
\mathcal{\tilde{O}}_p\sbr{\frac{1}{\sqrt{n}}\mbr{\alpha+\frac{\sqrt{p}}{\sqrt{m}}}+\frac{1}{n}}  \mathrm{for~ option~ I ~~and}~~
\mathcal{\tilde{O}}_p\sbr{\frac{p}{\sqrt{n}}\mbr{\alpha+\frac{1}{\sqrt{m}}}} \mathrm{for~ option~ II,}
$$
where $\frac{1}{\sqrt{n}}$ is the effective standard deviation for each machine with $n$ data points, $\alpha$ is the bias effect (price) of Byzantine machines, $\frac{1}{\sqrt{m}}$ is the averaging effect of $m$ normal machines, and $\frac{1}{n}$ is due to the dependence of the median on the skewness of the gradients. For strongly convex problems,   \cite{YinChenRB2018} proved that no algorithm can achieve an error lower than $\tilde{\Omega}\sbr{\frac{\alpha}{\sqrt{n}}+\frac{1}{\sqrt{nm}}}$ under regular conditions. Hence, this shows  the optimality of our methods  in some senses. As an natural extension of our first algorithm, our 2nd algorithm embeds the proximal algorithm \citep{ParikhBoyd2014} into the distributed procedure. They still perform well even under extremely mild conditions. Particularly,  it is more suitable for solving  very large scale or high-dimensional problems. In addition, algorithmic convergence  can be proved under more mild conditions, without requiring good initialization or a large sample size on each worker machine.

The remainder of this paper is organized as follows. In Section \ref{ProbForm}, we introduce the problem setup and communication-efficient surrogate likelihood framework for the distributed learning. Section \ref{sec-BCSL} proposes a Byzantine-robust CSL distributed learning algorithm and gives statistical guarantees under general conditions. Section \ref{sec-BCSLp} presents another Byzantine-robust CSL-proximal distributed learning algorithm and analyzes their theoretical properties. Section \ref{sec-NumericalExp} provides simulated and real data examples that illustrate the numerical performance of our algorithms, and thus validate the theoretical results.

{\bf Notations.}
For any positive integer $n$, we denote the set $\{1,2,\cdots,n\}$ by $[n]$ for brevity. For a vector, the standard $\ell_2$-norm and the $\ell_\infty$-norm is written by $\|\cdot\|_2$ and $\|\cdot\|_\infty$, respectively. For a matrix, the operator norm and the Frobenius norm is written by $\|\cdot\|_2$ and $\|\cdot\|_F$, respectively. For a different function $r: \mathbb{R}^p\rightarrow\mathbb{R}$, denote its partial derivative (or sub-differential set) with respect to the $k$-th argument by $\partial_kr$. Given a Euclidean space $\mathbb{R}^p$, $\vtheta_1,\vtheta_2\in \mathbb{R}^p$ and $r>0$, define $B(\vtheta_0,r)=\{\vtheta_1\in \mathbb{R}^p: \|\vtheta_0-\vtheta_1\|_2\leq r\}$ to be a closed ball with the center $\vtheta_0$, where $\theta_{1j}$ refers to the $j$th component of $\vtheta_{1}$. We assume that $\vtheta\in \vc\Theta\subset\mathbb{R}^p$ and $\vc\Theta$ is a convex and compact set  with diameter $D$. Let $\mathfrak{B}$ to be the set of Byzantine machines,  $\mathfrak{B}\subset\mathds{B}=\{2,\cdots,m+1\}$. Without loss of generality, we  assume that the 1st worker machine is normal and the other worker machine may be Byzantine. For matrices $A$ and $B$,  $A\succ B$ means $A-B$ is strictly positive. Given two sequences $\{a_n\}$ and $\{b_n\}$, we denote $a_n=\mathcal{O}(b_n)$ if $a_n\leq C_1b_n$ for some absolute positive constant $C_1$, $a_n=\Omega(b_n)$ if $a_n\geq C_2b_n$ for some absolute positive constant $C_2$. Furthermore, we also use notations  $\mathcal{\tilde{O}}(\cdot)$ and $\tilde{\Omega}(\cdot)$ to  hide  logarithmic factors in $\mathcal{O}(\cdot)$ and $\Omega(\cdot)$ respectively.

\section{Problem Formulation}\label{ProbForm}
In this section, we formally describe the problems setup. We focus on a standard statistical learning problem of (regularized) empirical risk minimization (ERM). In a distributed setting, suppose that we have access to one master and $m$ worker machines, and each worker machine independently communicates to the master one; each machine contains $n$ data points; and $\alpha m$ of the $m$ worker machines are Byzantine for some proportional level $\alpha<1/2$ and the remaining $1-\alpha$ fraction of worker machines are normal. Byzantine works can send any arbitrary values to the master machine. In addition, Byzantine workers may completely know the learning algorithm and are allowed to collude with each other \citep{YinChenRB2018}. In this setting, the total number of  data points is $N:=(m+1)n$.

Suppose that the observed data are sampled independently from an unknown probability distribution $\mathcal{D}$ over some metric space $\mathcal{Z}$. Let $\ell(\vtheta;\vZ)$ be a loss function of a parameter $\vtheta\in \vc\Theta\subset \mathbb{R}^p$ associated with the data point $\vZ$.  To measure the population loss of $\vtheta$, we define the expected risk by $F(\vtheta)=\mathbb{E}_{\vZ\sim \mathcal{D}}\ell(\vtheta;\vZ)$. Theoretically,
the true data-generating parameter we care about is a global minimizer of the population risk,
$$\vtheta^*=\mathrm{argmin}_{\vtheta\in \Theta}F(\vtheta).$$
 It is known that  negative log-likelihood functions are viewed as typical examples of the loss function $\ell(\cdot)$, for example, the Gaussian distribution corresponds to the least square loss, while the Bernoulli distribution for the logistic loss. Given that all the available samples $\{\vZ_i: i=1,\cdots,N\}$ are  stored on $m+1$ machines, the empirical loss of the $k$th machine is given as $f_k(\vtheta)=\frac{1}{|\mathcal{I}_k|}\sum_{i\in \mathcal{I}_k}\ell(\vtheta;\vZ_i)$, where $\mathcal{I}_k$ is the index set of samples over the $k$th machine with $|\mathcal{I}_k|=n=N/(m+1)$ for all $k\in[m+1]$, and $\mathcal{I}_j\cap \mathcal{I}_k=\emptyset$ for any $j\neq k$. In this paper, we are mainly concerned with learning $\vtheta^*$ via  minimizing the regularized empirical risk
\be\label{eq2-1}
\hat{\vtheta}=\mathrm{argmin}_{\vtheta\in \Theta}\{f(\vtheta)+g(\vtheta)\},
\ee
where
$f(\vtheta)=\frac{1}{m+1}\sum_{k=1}^{m+1}f_k(\vtheta),$
and $g(\vtheta)$ is a deterministic penalty function and independent of sample points,  such as the square $\ell_2$-norm in ridge estimation, the $\ell_1$-norm in the Lasso penalty \citep{Tibshirani1996}.

In the ideal non-Byzantine failure situation, one of core goals in the distributed framework is to develop   efficient distributed algorithms to approximate $\hat{\vtheta}$ well.  As a leading work in the literature,
\cite{JordanLeeYang2019} and \cite{WangKolarSZ2017} independently proposed an efficient distributed approach via the quasi-likelihood estimation. We now introduce the formulation of this method. Without loss of generality, we take the first machine as our master one. An initial estimator $\tilde{\vtheta}_0$ in the 1st machine is broadcasted to all other machines, which  compute their individual gradients at $\tilde{\vtheta}_0$. Then each gradient vector $\nabla f_k(\tilde{\vtheta}_0)$ is communicated back to the 1st machine. This constitutes one round of communication with a communication cost of $\mathcal{O}(mp)$. At the $(t+1)$-th iteration, the 1st machine calculates the following regularized surrogate loss
\be\label{eq2-2}
\tilde{\vtheta}_{t+1}=\mathrm{argmin}_\vtheta ~f_1(\vtheta)-\Big\langle \nabla f_1(\tilde{\vtheta}_{t})-\frac{1}{m+1}\sum_{k\in[m+1]}\nabla f_k(\tilde{\vtheta}_{t}),\vtheta\Big\rangle+g(\vtheta).
\ee
Next, the (approximate) minimizer $\tilde{\vtheta}_{t+1}$ without any aggregation operation is communicated to all the local machines, which is used to compute the local gradients, and then iterates as (\ref{eq2-2}) until convergence.

 Different from any first-order distributed optimization, the refined objective (\ref{eq2-2}) leverages both global first-order information and local higher-order information \citep{WangKolarSZ2017}. The idea of using such an adaptive enhanced function also has been developed in \cite{Shamiretal2014} and \cite{FanGuoWang2019}.

 Throughout the paper, we assume that $\ell()$ and $g()$ are convex in $\vtheta$, and $\ell()$ is twice continuously differentiable in $\vtheta$. We allow $g()$ to be non-smooth,
 for example, the $\ell_1$-penalty ($\lambda\|\vtheta\|_1$).

From (\ref{eq2-2}), we observe that this update for interested parameters strongly depends upon local gradients at any iteration. Hence,   the standard learning algorithm only based on average aggregation of the workers' messages would be arbitrarily skewed if some of local workers are Byzantine-faulty machines. To address this robust-related problem, we develop  two Byzantine-robust distributed learning algorithms given in next two sections.

\section{Byzantine-robust CSL distributed learning}\label{sec-BCSL}

In this section, we introduce our first communication-efficient Byzantine-robust distributed learning algorithm based on the CSL framework, and particularly introduce two robust operations to handle the Byzantine failures. After giving some technical assumptions, we  present optimization error and statistical analysis of multi-step estimators. In the end of this section,  we further clarify our results by a concrete example of   generalized linear models (GLMs).

\subsection{Byzantine-robust CSL distributed algorithm}

When the Byzantine failures occur, the aggregation rule  (\ref{eq2-2}) will be sensitive to the bad gradient values. More precisely, although the master machine communicates with the worker machines via some predefined protocol,  the Byzantine machines do not have to obey this protocol and may send arbitrary messages to the master machine. At this time, the gradients $\{\nabla f_k(\cdot): k=2,\cdots,m+1\}$ received by the master machine are not always reliable, since the information from Byzantine machine may be completely out of its local data. To state it clearly, we assume that the Byzantine workers can provide arbitrary values written by the symbol ``$*$" to the master machine. In this situation, several robust operations should be implemented to substitute the simply average of local gradients as in (\ref{eq2-2}).

Inspired by  robust techniques developed recently in \cite{YinChenRB2018}, we apply for the coordinate-wise median and coordinate-wise trimmed mean  to formulate our Byzantine-robust CSL distributed learning algorithm.
\begin{defi} (Coordinate-wise median)
For vectors $\vc x^i\in \mathbb{R}^p$, $i\in[m]$, the coordinate-wise median $\vc g=\mathbf{med}\{\vc x^i: i\in [m]\}$ is a vector with its $k$-th coordinate being $g_k=\mathbf{med}\{x_k^i: i\in[m]\}$ for each $k\in[p]$, where $\mathbf{med}$ is the usual (one-dimensional) median.
\end{defi}
\begin{defi} (Coordinate-wise trimmed mean)
For $\beta\in [0, \frac{1}{2})$ and vectors $\vc x^i\in \mathbb{R}^p$, $i\in[m]$, the coordinate-wise $\beta$-trimmed mean $\vc g=\mathbf{trmean}_\beta\{\vc x^i: i\in[m]\}$ is a vector with its $k$-th coordinate being $g_k=\frac{1}{(1-2\beta)m}\sum_{x\in U_k}x$ for each $k\in [p]$. Here $U_k$ is a subset of $\{x_k^1,\cdots,x_k^m\}$ obtained by removing the largest and small $\beta$ fraction of its elements.
\end{defi}
See Algorithm 1 below for details, and we call it Algorithm BCSL.
In each parallel iteration of Algorithm \ref{alg1}, the 1st machine (the normal master machine) broadcasts the current model parameter to all worker machines. The normal worker machines calculate their own gradients of loss functions based on their local data and then send them to the 1st machine. Considering that the Byzantine machines may send any messages due to their abnormal or adversarial behavior,  we implement the coordinate-wise median or trimmed mean operation for these received gradients at the 1st machine. Then
the aggregation  algorithm in \eqref{eq-alg1} is conducted to update the  global parameter.

\begin{algorithm}[H]\label{alg1}
        \caption{Byzantine-Robust CSL distributed learning (BCSL)}
        \KwIn{Initialize estimator $\vtheta_0$, algorithm parameter $\beta$ (for Option II) and number of iteration $T$}

        \For{$t=0,1,2,\cdots,T-1$}{
            \emph{The 1st machine}: send $\vtheta_t$ to other worker machines. \\
            \For{$\mathbf{all}~k\in\{2,3,\cdots,m+1\}~\mathbf{parallel}$}
                {\emph{Worker machine $k$}: evaluate local gradient\\
                \be
                \vc h_k(\vtheta_t)\leftarrow
                \left\{\begin{array}{cl}
                \nabla f_k(\vtheta_t) & \mathrm{normal~ worker ~ machines,} \\
                * & \mathrm{Byzantine~ machines},                                        \end{array}
                \right.\nonumber
                \ee\\
                send $\vc h_k(\vtheta_t)$ to the 1st machine.
            }
        \emph{The 1st machine}: evaluates aggregate gradients\\
            \be
            \vc h(\vtheta_t)\leftarrow
                \left\{\begin{array}{ll}
                \mathbf{med}\{\vc h_k(\vtheta_t): k\in [m+1]\}& \mathrm{Option~ I}, \\
                \mathbf{trmean}_\beta\{\vc h_k(\vtheta_t): k\in [m+1]\} & \mathrm{Option~ II},                                        \end{array}
                \right.\nonumber
            \ee computes\\
         \be\label{eq-alg1}
         \vtheta_{t+1}=\mathrm{argmin}_\vtheta\left\{f_1(\vtheta)-\langle\vc \nabla f_1(\vtheta_t)-\vc h(\vtheta_t),\vtheta\rangle+g(\vtheta)\right\},
         \ee
         and then broadcasts it to local machines.
        }
    \KwOut{$\vtheta_T$.}
\end{algorithm}

\vskip 8pt
In order to  provide an optimization error and statistical error of Algorithm 1, we need to introduce some basic conditions  for our theoretical analysis.
\begin{ass}\label{ass-3-1}
(Lipschitz Conditions and Smoothness). For any $\vZ\in \mathcal{Z}$ and  $k\in[p]$, the partial derivative $\partial_k\ell(\cdot;\vZ)$ with respect to its first component is $L_k$-Lipschitz.The loss function  $\ell(\cdot;\vZ)$ itself is $L$-smooth in sense that its  gradient vector is $L$-Lipschitz continuous under the $\ell_2$-norm. {Let $\tilde{L}=\sqrt{\sum_{k=1}^pL_k^2}$. Further assume that the population loss function $F(\cdot)$ is $L_F$-smooth.}
\end{ass}

For Option I in Algorithm 1: median-based algorithm, some moment conditions of the gradient of $\ell()$ is introduced to control stochastic behaviors. 

\begin{ass}\label{ass-3-2}
There exist two constants $V$ and $S$, for any $\vtheta\in\vc\Theta$ and all $\vZ \in \mathcal{Z}$, such that

(i) (Bounded variance of gradient). $V^2{\bf I}\succ Var(\nabla \ell(\vtheta;\vZ))$.

(ii) (Bounded skewness of gradient). $\|\gamma(\nabla\ell(\vtheta;\vZ))\|_\infty\leq S$.
Here $\gamma(\cdot)$ refers to the coordinate-wise skewness of vector-valued random variables.
\end{ass}
Assumption \ref{ass-3-2} is standard in the literature and is  satisfied in many learning problems.  See Proposition 1 in \cite{YinChenRB2018} for a specific linear regression problem.
\begin{ass}\label{ass-3-3}
(Strong convexity). $f+g$ has a unique minimizer $\hat{\vtheta}\in \mathbb{R}^p$, and is $\rho$-strongly convex in $B(\hat{\vtheta},R)$ for some $R>0$ and $\rho>0$.
\end{ass}
\begin{ass}\label{ass-3-4}
(Homogeneity) $\|\nabla^2f_1(\vtheta)-\nabla^2F(\vtheta)\|_2\leq \delta$ for $\delta\geq0$ and $\vtheta\in B(\hat{\vtheta},R)$.
\end{ass}

In most existing studies, it is usually assumed that the population risk $F$ is smooth and strongly convex. The empirical risk $f$ also enjoys such good properties, as long as $\{\vc z_i: i=1,\cdots,N\}$ are i.i.d. and the total sample size $N$ is sufficiently large relative to $p$ \citep{FanGuoWang2019}. From (\ref{eq-alg1}) in Algorithm \ref{alg1}, we know the local data at the 1st  machine are used to optimize. So we need to control the gap between $\nabla^2f_1(\vtheta)$ and $\nabla^2F(\vtheta)$ to contract optimization rate of the algorithm. The similarity between $f_1$ and $F$ is depicted by Assumption \ref{ass-3-4}.
Indeed, the empirical risk $f_1$ should not be too far away from their population risk $F$ as long as the sample size $n$ of the 1st  machine is not too small. \cite{Meietal2018} showed it holds with reasonably small $\delta$ and large $R$ with high probability under general conditions. Specially, a large $n$ implies a small homogeneity index $\delta$. Obviously, Assumption \ref{ass-3-4} always holds when taking $\delta=\sup_{\vtheta\in B(\hat{\vtheta},R)}\sbr{\|\nabla^2f_1(\vtheta)\|_2+\|\nabla^2F(\vtheta)\|_2}$.

The following theorem establishes the global convergence of the proposed algorithm BCSL, involving a trade off between the optimization error  and statistical error.
\begin{thm}\label{thm1}
Assume that Assumptions \ref{ass-3-1}-\ref{ass-3-4} hold, the iterates $\{\vtheta_t\}_{t=0}^\infty$ produced by Option I in Algorithm \ref{alg1} with $\vtheta_0\in B(\hat{\vtheta}, R)$ for $t\geq0$, $\rho>\delta+2\Delta_{nm\alpha}/R$, and the fraction $\alpha$ of Byzantine machines satisfies
\be\label{eq3-0}
\alpha+\sqrt{\frac{p\log(1+n(m+1)\tilde{L}D)}{m(1-\alpha)}}+0.4748\frac{S}{\sqrt{n}}\leq \frac{1}{2}-\varepsilon
\ee
for some $\varepsilon>0$.
Then, with probability at least $1-\frac{4p}{(1+n(m+1)\tilde{L}D)^p}$, we have
$$\|\vtheta_{t+1}-\hat{\vtheta}\|_2\leq \frac{\delta}{\rho}\|\vtheta_{t}-\hat{\vtheta}\|_2+\frac{2}{\rho}\Delta_{nm\alpha},$$
where
$$\Delta_{nm\alpha}=\frac{2\sqrt{2}}{(m+1)n}+\frac{\sqrt{2}C_\varepsilon V}{\sqrt{n}}\sbr{\alpha+\sqrt{\frac{p\log(1+n(m+1)\tilde{L}D)}{m(1-\alpha)}}+0.4748\frac{S}{\sqrt{n}}}.$$
Here  $\tilde{L}=\sqrt{\sum_{k=1}^pL_k^2}$ and
$C_\varepsilon=\sqrt{2\pi}\exp\sbr{\frac{1}{2}(\Phi^{-1}(1-\varepsilon))^2}$,
where $\Phi^{-1}(\cdot)$ being the inverse of the cumulative distribution function of the standard Gaussian distribution $\Phi(\cdot)$.
\end{thm}


Theorem \ref{thm1} shows the linear convergence of $\vtheta_t$, which depends explicitly on the homogeneity index $\delta$,  the strong convex index $\rho$, and the fraction of Byzantine machines $\alpha$. The result is viewed as an extension of that in \cite{JordanLeeYang2019} and \cite{FanGuoWang2019}. A significant difference from theirs is that, we allow the initial estimator to be inaccurate, and with high probability we have more explicit rates of convergence on optimization errors under the Byzantine failures.

Specially, the factor $C_\varepsilon$  in Theorem \ref{thm1}  is a function of $\varepsilon$, for example, $C_\varepsilon\approx4$ if setting $\varepsilon=1/6$. After running $T$ for Algorithm \ref{alg1}, with high probability, we have
\be\label{eq3-T1}
\|\vtheta_T-\hat{\vtheta}\|_2\leq \sbr{\frac{\delta}{\rho}}^T\|\vtheta_{0}-\hat{\vtheta}\|_2+\frac{2}{\rho-\delta}\Delta_{nm\alpha}.
\ee
Notice that $\log (1-x)\leq -x$. Theorem \ref{thm1} guarantees that after parallel iterating $T\geq \frac{\rho}{\rho-\delta}\log \frac{(\rho-\delta)\|\vtheta_{0}-\hat{\vtheta}\|_2}{2\Delta_{nm\alpha}}$, with high probability we can obtain a solution $\tilde{\vtheta}=\vtheta_T$ with an error
$$\|\tilde{\vtheta}-\hat{\vtheta}\|_2\leq \frac{4}{\rho-\delta}\Delta_{nm\alpha}.$$
In this case, the derived rate of the statistical error between $\tilde{\vtheta}$ and the centralized empirical
risk minimizer $\hat{\vtheta}$ is of the order $\mathcal{O}\sbr{\frac{\alpha}{\sqrt{n}}+\sqrt{\frac{p\log (nm)}{nm}}+\frac{1}{n}}$ up to some constants, alternatively, $\tilde{\mathcal{O}}\sbr{\frac{\alpha}{\sqrt{n}}+\frac{1}{\sqrt{nm}}+\frac{1}{n}}$ up to the logarithmic factor. Note that
$$\tilde{\mathcal{O}}\sbr{\frac{\alpha}{\sqrt{n}}+\frac{1}{\sqrt{nm}}+\frac{1}{n}}=\tilde{\mathcal{O}}\sbr{\frac{1}{\sqrt{n}}(\alpha+\frac{1}{\sqrt{m}})+\frac{1}{n}}.$$
Intuitively, the above error rate is a near optimal rate that one should target, as $\frac{1}{\sqrt{n}}$ is the effective standard deviation for each machine with $n$ data points, $\alpha$ is the bias effect of Byzantine machines, $\frac{1}{\sqrt{m}}$ is the averaging effect of $m$ normal machines, and $\frac{1}{n}$ is the effect of the dependence of median on skewness of the gradients. If $n\gtrsim m$, then $\tilde{\mathcal{O}}\sbr{\frac{\alpha}{\sqrt{n}}+\frac{1}{\sqrt{nm}}+\frac{1}{n}}
=\tilde{\mathcal{O}}\sbr{\frac{\alpha}{\sqrt{n}}+\frac{1}{\sqrt{nm}}}$ is the order-optimal rate \citep{YinChenRB2018}.
When $\alpha=0$ (no Byzantine machine), one sees the usual scaling $\frac{1}{\sqrt{nm}}$ with the global sample size; when $\alpha\neq0$ (some machines are Byzantine), their influence remains bounded and is proportional to $\alpha$. So we do not sacrifice the quality of learning  to guard against Byzantine failures, provided that the Byzentine failure proportion satisfies  $\alpha=\mathcal{O}(1/\sqrt{m})$ .

Remark also that, our results for convex problems follow for any finite-bounded  $R$ of the initial radius.

We next turn to an analysis for Option II in Algorithm \ref{alg1}:  The robust distributed learning based on coordinate-wise trimmed mean. Compared to Option I, a stronger assumption on the tail behavior of the partial derivatives of the loss functions is needed as follows.
\begin{ass}\label{ass-3-5}
(Sub-exponential gradients). For all $k\in[p]$ and $\vtheta\in \vc \Theta$, the partial derivative of $\ell(\vtheta;\vZ)$ with respect to the $k$-th coordinate of $\vtheta$, $\partial_k\ell(\vtheta;\vZ)$, is $\upsilon$-sub-exponential.
\end{ass}
The sub-exponential assumption implies that all the moments of the derivatives are bounded. Hence, this  condition is stronger a little than the bounded absolute skewness (Assumption \ref{ass-3-2}(ii)). Fortunately, {Assumption \ref{ass-3-5} can be satisfied in some learning problems, and see Proposition 2 in \cite{YinChenRB2018}.}

\begin{thm}\label{thm2}
Assume that Assumptions \ref{ass-3-1}, \ref{ass-3-3}-\ref{ass-3-5} hold, the iterates $\{\vtheta_t\}_{t=0}^\infty$ produced by Option II in Algorithm \ref{alg1} with $\vtheta_0\in B(\hat{\vtheta}, R)$ for $t\geq0$, $\rho>\delta+2\Delta_{nm\beta}/R$, and $\alpha\leq \beta\leq \frac{1}{2}-\varepsilon$ for some $\varepsilon>0$. Then, with probability at least $1-\frac{2p(m+2)}{(1+n(m+1)\tilde{L}D)^p}$, we have
$$\|\vtheta_{t+1}-\hat{\vtheta}\|_2\leq \frac{\delta}{\rho}\|\vtheta_{t}-\hat{\vtheta}\|_2+\frac{2}{\rho}\Delta_{nm\beta},$$
where
$$\Delta_{nm\beta}=\sbr{\frac{\upsilon p}{\varepsilon}\mbr{\frac{3\sqrt{2}\beta}{\sqrt{n}}+\frac{2}{\sqrt{nm}}}\sqrt{\log(1+n(m+1)\tilde{L}D)+\frac{\log (1+m)}{p}}+\mathcal{\tilde{O}}\sbr{\frac{\beta}{n}+\frac{1}{nm}}}.
$$
\end{thm}

Theorem \ref{thm2} also shows the linear convergence of $\vtheta_t$, which depends explicitly on the homogeneity index $\delta$, the strong convex index $\rho$, and the trimmed mean index $\beta$ with choosing the index to satisfy $\beta\geq\alpha$, where $\alpha$ is a
fraction of Byzantine machines. Note that, the hyperparameter $\upsilon$ in Assumption \eqref{ass-3-5} only affect the statistical error appearing in $\Delta_{nm\beta}$.

 Similar to (\ref{eq3-T1}), we also have
$$
\|\vtheta_T-\hat{\vtheta}\|_2\leq \sbr{\frac{\delta}{\rho}}^T\|\vtheta_{0}-\hat{\vtheta}\|_2+\frac{2}{\rho-\delta}\Delta_{nm\beta}
$$
after $T$-step iterations.
 By running $T\geq\frac{\rho}{\rho-\delta}\log \frac{(\rho-\delta)\|\vtheta_{0}-\hat{\vtheta}\|_2}{2\Delta_{nm\beta}}$ parallel computations, we can obtain a solution $\tilde{\tilde{\vtheta}}=\vtheta_T$ satisfying $\|\tilde{\tilde{\vtheta}}-\hat{\vtheta}\|_2=\mathcal{\tilde{O}}\sbr{\frac{\beta}{\sqrt{n}}+\frac{1}{\sqrt{nm}}}$,
 since the term $\Delta_{nm\beta}$ can be reduced to be
 $\Delta_{nm\beta}=\mathcal{O}\sbr{\frac{\upsilon p}{\varepsilon}\mbr{\frac{\beta}{\sqrt{n}}+\frac{1}{\sqrt{nm}}}\sqrt{\log(nm)}}$.

It should be pointed out that, the trimmed mean index $\beta$ is strictly controlled by the fraction of Byzantine machines $\alpha$, that is  $\frac{1}{2}-\varepsilon\geq\beta\geq\alpha$. By choosing $\beta=c\alpha$ with $c\geq1$, {we still achieve the optimization error rate $\mathcal{\tilde{O}}\sbr{\frac{\alpha}{\sqrt{n}}+\frac{1}{\sqrt{nm}}}$, which  is also  order-optimal in the statistical literature.}

We now take comparable analysis for the above two Byzantine-robust CSL distributed learning in  Algorithm 1 (Options I and II). The trimmed-mean-based algorithm (Option II) has an order-optimal optimization error rate $\mathcal{\tilde{O}}\sbr{\frac{\alpha}{\sqrt{n}}+\frac{1}{\sqrt{nm}}}$. By contrast, the median-based algorithm (Option I) has the rate $\tilde{\mathcal{O}}\sbr{\frac{\alpha}{\sqrt{n}}+\frac{1}{\sqrt{nm}}+\frac{1}{n}}$ involving an additional term $\frac{1}{n}$, and thus the optimality is achieved only for $n\succeq m$.
Note that Option I algorithm needs milder moment conditions (bounded skewness) on the tail of the loss derivatives than the Option II algorithm (sub-exponentiality). In other words, this provides a profound insight into the underlying relation between the tail decay of the loss derivatives and the block number of local machines ($m$).

On the other hand,  Algorithm 1 based on Option II has an additional parameter  $\beta$ such that $1/2>\beta\geq \alpha$, which requires that the fraction of Byzantine machines is absolutely dominated by the normal ones for guaranteeing robustness. In contrast to this, Algorithm 1 based on Option I  has a weaker restriction on $\alpha$.

\subsection{Specific Example: Generalized linear models}\label{subsec3-2}
We now unpack Theorems \ref{thm1} and \ref{thm2} in generalized linear models, taking into consideration the effects of iterations in the proposed  estimator, the roles of the initial estimator and the Byzantine failures. We will find an explicit rate of convergence of $\delta$ in Assumption \ref{ass-3-4} in the setting of generalized linear models. Theorem \ref{thm1} and \ref{thm2} guarantee that after running $T$ parallel iterations, with high probability we achieve an optimization error with a linear rate. Moreover, through finite steps, the optimization errors are eventually negligible in comparison with the statistical errors. We will give a specific analysis below.

For GLMs,  the loss function is the negative partial log-likelihood of an exponential-type variable of the response given any input feature $\vc x$. Suppose that the i.i.d. pairs $\{\vZ_i=(\vc x_i^T, y_i)^T\}_{i=1}^N$ are drawn from a generalized linear model. Recall that the conditional density of $y_i$ given $\vc x_i$ takes the form
$$h(y_i;\vc x_i,\vtheta^*)=c(\vc x_i,y_i)\exp\sbr{y_i\vc x_i^T\vtheta^*-b(\vc x_i^T\vtheta^*)},$$
where $b(\cdot)$ is some known convex function, and $c(\cdot)$ is a known function such that $h(\cdot)$ is a valid probability density function. The loss function corresponding to the negative log likelihood of the whole data is given by
$f(\vtheta)=\frac{1}{m+1}\sum_{k=1}^{m+1}f_k(\vtheta)$ with
\be\label{eq3-7}
f_k(\vtheta)=\frac{1}{n}\sum_{i\in\mathcal{I}_k}\ell(\vtheta;\vZ_i)
~~\mathrm{and}~~\ell(\vtheta;\vZ_i)=b(\vc x_i^T\vtheta)-y_i\vc x_i^T\vtheta.
\ee

We further impose the following standard regularity technical conditions.

\begin{ass}\label{ass-3-6}
(i) There exist universal positive constants $A_i (i=1,2,3)$ such that $A_1\leq \|\vc \Sigma\|_2\leq A_2p^{A_3}$, where $\vc \Sigma=E(\vc x_i\vc x_i^T)$.

(ii) $\{\vc \Sigma^{-1/2}\vc x_i\}_{i=1}^N$ are i.i.d. sub-Gaussian random vectors with bounded $\|\vc \Sigma^{-1/2}\vc x_i\|_{\psi_2}$.

(iii) $\{\vc \Sigma^{-1/2}\vc x_i\}_{i=1}^N$ are i.i.d. random vectors, and each component is $v$-sub-exponential.

(iv) For all $t\in \mathbb{R}$, $|b''(t)|$ and $|b'''(t)|$ are both bounded.

(v) $\|\vtheta^*\|_2$ is bounded.
\end{ass}

\begin{ass}\label{ass-3-7}
$F+g$ is $\rho$-strongly convex in $B(\vtheta^*,2R)$, where $R<A_4p^{A_5}$ for some universal constants $A_4$, $A_5$ and $\rho>0$.
\end{ass}

Assumptions (\ref{ass-3-6})(i)(ii)(iv)(v)-(\ref{ass-3-7}) have been proposed by \cite{FanGuoWang2019} to establish the rate of optimization errors for a distributed statistical inference. In our paper, these assumptions mainly are used to obtain asymptotic properties of $\vtheta_t$ in Option I algorithm. For establishing the rate of optimization errors of Option II algorithm, we need Assumption (\ref{ass-3-6}) {(iii) (sub-exponentiality) instead of Assumption (\ref{ass-3-6})(ii) (sub-Gaussian), which is similar to Theorem \ref{thm2}.}

From (\ref{eq3-7}), we easily obtain
$$\nabla f_1(\vtheta)=\frac{1}{n}\sum_{i\in\mathcal{I}_k}[b'(\vc x_i^T\vtheta)-y_i]\vc x_i ~~\mathrm{and}~~\nabla^2 f_1(\vtheta)=\frac{1}{n}\sum_{i\in\mathcal{I}_k}b''(\vc x_i^T\vtheta)\vc x_i\vc x_i^T.$$
By Lemma A.5 in \cite{FanGuoWang2019}, we immediately  get
$$\max_{\vtheta\in B(\hat{\vtheta},R)}\|\nabla^2 f_1(\vtheta)-\nabla^2 F(\vtheta)\|_2=\mathcal{O}_p\sbr{\|\vc\Sigma\|_2\sqrt{\frac{p(\log p+\log n)}{n}}},$$
as long as $n>cp$ for a given positive constant $c$. So, $\delta=O\sbr{\|\vc\Sigma\|_2\sqrt{p(\log p+\log n)/n}}$ can be chosen with high probability. From Theorems \ref{thm1} and \ref{thm2}, we have the contraction factor $O\sbr{\kappa\sqrt{p(\log p+\log n)/n}}$ with $\kappa=\|\vc\Sigma\|_2/\rho$. The explicit parameter $\kappa$ is comparable to the condition number in \cite{JordanLeeYang2019}, where  finite $p$ and $\kappa$ were imposed.

Equipped with these above facts, we have the following corollary.

\begin{thm}\label{thm3}
Assume that Assumptions \ref{ass-3-6}(i)(ii)(iv)(v) and \ref{ass-3-7} hold and with probability tending to one $\vtheta_0\in B(\hat{\vtheta},R)$ for some $R>\|\hat{\vtheta}-\vtheta^*\|_2$. The iterates $\{\vtheta_t\}_{t=0}^\infty$ produced by Option I in Algorithm \ref{alg1} for $t\geq0$, $\rho>\eta^{1/2}+2\Delta_{nm\alpha}/R$, and $\alpha$ satisfies (\ref{eq3-0}). Then, after $t$ parallel iterations, we have
$$\|\vc\vtheta_t-\hat{\vtheta}\|_2=\mathcal{O}_p\sbr{\eta^{t/2}\|\vtheta_0-\hat{\vtheta}\|+\frac{2}{\rho-\eta^{1/2}}\Delta_{nm\alpha}}, \forall \,t\geq0,$$
where $\eta=\kappa^2p(\log p+\log n)/n$ and $\Delta_{nm\alpha}$ is defined Theorem \ref{thm1}.
\end{thm}

\begin{thm}\label{thm4}
Assume that Assumptions \ref{ass-3-6}(i)(iii)(iv)(v) and \ref{ass-3-7} hold, $\ell(\vtheta;\vZ)$ satisfies Assumption \ref{ass-3-5}, and with probability tending to one $\vtheta_0\in B(\hat{\vtheta},R)$ for some $R>\|\hat{\vtheta}-\vtheta^*\|_2$. The iterates $\{\vtheta_t\}_{t=0}^\infty$ produced by Option II in Algorithm \ref{alg1} for $t\geq0$, $\rho>\eta^{1/2}+2\Delta_{nm\beta}/R$, and $\alpha\leq\beta\leq \frac{1}{2}-\varepsilon$ for some $\varepsilon>0$. Then, after $t$ parallel iterations, we have
$$\|\vc\vtheta_t-\hat{\vtheta}\|_2=\mathcal{O}_p\sbr{\eta^{t/2}\|\vtheta_0-\hat{\vtheta}\|+\frac{2}{\rho-\eta^{1/2}}\Delta_{nm\beta}}, \forall\, t\geq0,$$
where $\eta=\kappa^2p(\log p+\log n)/n$ and $\Delta_{nm\beta}$ is defined Theorem \ref{thm2}.
\end{thm}

Theorems \ref{thm3} and \ref{thm4} clearly present how Algorithm \ref{alg1} depend on structural parameters of the problem. When $\kappa$ is bounded, $\eta=\mathcal{O}(p(\log p+\log n)/n)$, through finite steps, $\eta^{t/2}$ can be much smaller than $\Delta_{nm\alpha} (\Delta_{nm\beta})$. Thus,
\be\label{eq3-5-1}
\|\vc\vtheta_t-\hat{\vtheta}\|_2=E_n
:=\left\{\begin{array}{cl}
         \mathcal{O}_p\sbr{\frac{\alpha}{\sqrt{n}}+\sqrt{\frac{p\log (nm)}{nm}}+\frac{1}{n}},  & \mathrm{Option~ I;} \\
          \mathcal{O}_p\sbr{p\mbr{\frac{\beta}{\sqrt{n}}+\frac{1}{\sqrt{nm}}}\sqrt{\log(nm)}}, & \mathrm{Option~ II.}
        \end{array}
 \right.
\ee
So, By contrast to that results in \cite{JordanLeeYang2019}, we allow an inaccurate initial value $\vtheta_0$ and give more explicit rates of optimization error even when $p$ diverges.

We know that the statistical error of the estimator $\vtheta_t$ can be controlled by the optimization error of $\vtheta_t$ and statistical error of $\hat{\vtheta}_t$, that is,
\be\label{eq3-5-2}
\|\vtheta_t-\vtheta^*\|_2\leq \|\vtheta_t-\hat{\vtheta}\|_2+\|\hat{\vtheta}-\vtheta^*\|_2.
\ee
Note that the first term is not a deterministic optimization error, it holds in probability. Therefore, it is an optimization error in a statistical sense. We call it statistical optimization error.
The second term is of order $\mathcal{O}_p(\sqrt{p/(nm)})$ under mild conditions, which has been well studied in statistics. Thus, the statistical error of $\vtheta_t$ is controlled by the magnitude of the first term. If we adopt two-step iteration, and $\|\vtheta_0-\vtheta^*\|_2=\mathcal{O}_p(\sqrt{p/n})$ when $\vtheta_0$ is obtained in 1st machine, one gets
$$\|\vtheta_t-\vtheta^*\|_2=E_n$$
by (\ref{eq3-5-1})-(\ref{eq3-5-2}), where $E_n$ is defined in (\ref{eq3-5-1}),
provided that for Option I
$$n\gg \mbr{\alpha^{-1}p^{3/2}+p^{2/3}N^{1/3}\log^{-1}N+p^3}\log N=p^3(\alpha^{-1}p^{1/2}+1)\log N+(p^2N)^{1/3},$$
 or for Option II
$$n\gg \mbr{\beta^{-1}\sqrt{p\log N}+(Np\log N)^{1/3}}.$$
{It implies that Algorithm \ref{alg1} has the order-optimal statistical error rate, for Option I which needs $n\gtrsim m$.}


\section{Byzantine-robust CSL-proximal distributed learning}\label{sec-BCSLp}

For Algorithm \ref{alg1}, we establish the contraction rates of optimization errors and statistical analysis under sufficiently strong convexity of $f_1+g$ and small discrepancy between $\nabla^2f_1$ and $\nabla^2 F$. It requires the data points of each machine to be large enough, and even the required data size $n$ of each machine depends on structural parameters, which may not be realistic in practice. The coordinate-wise median and coordinate-wise trimmed mean are proposed in algorithm \ref{alg1}, which is robust for the Byzantine failure, but it is unstable in the optimization process of the 1st worker machine even for moderate $n$. In the section, we propose another Byzantine-robust CSL algorithm via embedding the proximal algorithm. See \cite{Rockafellar1976} and \cite{ParikhBoyd2014} for proximal algorithm.

\subsection{Byzantine-robust CSL-proximal distributed algorithm}

First, recall that the proximal operator $\mathrm{prox}_h: \mathbb{R}^p\rightarrow\mathbb{R}^p$ is defined by
$$\mathrm{prox}_h(\vc v)=\mathrm{argmin}_x\sbr{h(\vc x)+\frac{1}{2}\|\vc x-\vc v\|_2^2}.$$
By the proximal operator of the function $\lambda^{-1}h$ with $\lambda>0$, the proximal algorithm for minimizing $h$ iteratively computes
$$\vc x_{t+1}=\mathrm{prox}_{\lambda^{-1}h}(\vc x_t)=\mathrm{argmin}_{\vc x\in \mathbb{R}^p}\sbr{h(\vc x)+\frac{\lambda}{2}\|\vc x-\vc x_t\|_2^2}$$
starting from some initial value $\vc x_0$. \cite{Rockafellar1976} showed the $\{\vc x_t\}_{t=0}^\infty$ converges linearly to some $\hat{\vc x}\in \mathrm{argmin}_{\mathbb{R}^p}h(\vc x)$.

For our problem (\ref{eq2-1}), the proximal iteration algorithm is
$$\vtheta_{t+1}=\mathrm{prox}_{\lambda^{-1}(f+g)}(\vtheta_t)=\mathrm{argmin_{\vtheta\in \mathbb{R}^p}}\sbr{f(\vtheta)+g(\vtheta)+\frac{\lambda}{2}\|\vtheta-\vtheta_t\|_2^2}.$$
In our setting, we adopt the distributed learning, and optimization is mainly on the 1st worker machine. The $g(\cdot)$ is a penalty function, which is used to the optimization step, and keep the local data of the 1st worker machine in $f_1(\cdot)$. So, the penalty function in our proximal algorithm becomes $g(\vtheta)+\frac{\lambda}{2}\|\vtheta-\vtheta_t\|_2^2$. Further, the optimization (\ref{eq-alg1}) in Algorithm \ref{alg1} is replaced by
\be\label{eq4-1}
\vtheta_{t+1}=\mathrm{argmin}_\vtheta\left\{f_1(\vtheta)-\langle\vc \nabla f_1(\vtheta_t)-\vc h(\vtheta_t),\vtheta\rangle+g(\vtheta)+\frac{\lambda}{2}\|\vtheta-\vtheta_t\|_2^2\right\},
\ee
where $h(\vtheta_t)$ is the gradient information at $\vtheta_t$ from the other worker machines. This optimization (\ref{eq4-1}) can make the Byzantine-robust CSL-proximal distributed learning converges rapidly. See the following Algorithm \ref{alg2}. We call it Algorithm BCSLp.

\begin{algorithm}[H]\label{alg2}
        \caption{Byzantine-robust CSL-proximal distributed learning (BCSLp)}
        \KwIn{Initialize estimator $\vtheta_0$, algorithm parameter $\beta$ (for Option II) and number of iteration $T$}

        \For{$t=0,1,2,\cdots,T-1$}{
            \emph{The 1st machine}: send $\vtheta_t$ to other worker machines. \\
            \For{$\mathbf{all}~k\in\{2,3,\cdots,m+1\}~\mathbf{parallel}$}
                {\emph{Worker machine $k$}: evaluate local gradient\\
                \be
                \vc h_k(\vtheta_t)\leftarrow
                \left\{\begin{array}{cl}
                \nabla f_k(\vtheta_t) & \mathrm{normal~ worker ~ machines,} \\
                * & \mathrm{Byzantine~ machines},                                        \end{array}
                \right.\nonumber
                \ee\\
                send $\vc h_k(\vtheta_t)$ to the 1st machine.
            }
        \emph{The 1st machine}: evaluates aggregate gradients\\
            \be\label{eq-alg2}
            \vc h(\vtheta_t)\leftarrow
                \left\{\begin{array}{ll}
                \mathbf{med}\{\vc h_k(\vtheta_t): k\in [m+1]\}& \mathrm{Option~ I}, \\
                \mathbf{trmean}_\beta\{\vc h_k(\vtheta_t): k\in [m+1]\} & \mathrm{Option~ II},                                        \end{array}
                \right.
            \ee computes\\
         \be
         \vtheta_{t+1}=\mathrm{argmin}_\vtheta\left\{f_1(\vtheta)-\langle\vc \nabla f_1(\vtheta_t)-\vc h(\vtheta_t),\vtheta\rangle+g(\vtheta)+\frac{\lambda}{2}\|\vtheta-\vtheta_t\|_2^2\right\},\nonumber
         \ee
         and then broadcasts to other machines.
        }
    \KwOut{$\vtheta_T$.}
\end{algorithm}
The above Algorithm \ref{alg2} is a communication-efficient Byzantine-robust accurate statistical learning, which adopts coordinate-wise median and coordinate-wise trimmed mean cope with Byzantine fails, and use the proximal algorithm as the backbone. In each iteration, it has one round of communication and one optimization step similar to Algorithm \ref{alg1}. It is regularized version of Algorithm \ref{alg1} by adding a strict convex quadratic term in the objective function. The technique has been used in the distributed stochastic optimization such as accelerating first-order algorithm \citep{LeeLinMY2017} and regularizing sizes of updates \citep{WangWangS2017}, and in the communication-efficient accurate distributed statistical estimation \citep{FanGuoWang2019}.

Now, we give contraction guarantees for Algorithm \ref{alg2}.

\begin{thm}\label{thm5}
Assume that Assumptions \ref{ass-3-1}-\ref{ass-3-4} hold, the iterates $\{\vtheta_t\}_{t=0}^\infty$ produced by Option I in Algorithm \ref{alg2} with $\vtheta_0\in B(\hat{\vtheta}, R/2)$ for $t\geq0$, $\sbr{\frac{\delta+2R^{-1}\Delta_{nm\alpha}}{\rho+\lambda}}^2<\frac{\rho}{\rho+2\lambda}$, and the fraction $\alpha$ of Byzantine machines satisfies (\ref{eq3-0}).
Then, with probability at least $1-\frac{4p}{(1+n(m+1)\tilde{L}D)^p}$, we have
\be\label{eq4-2}
\|\vtheta_{t+1}-\hat{\vtheta}\|_2\leq \frac{\frac{\delta}{\rho+\lambda}\sqrt{\rho^2+2\lambda\rho}+\lambda}{\rho+\lambda}\|\vtheta_{t}-\hat{\vtheta}\|_2+\frac{2}{\rho+\lambda}\Delta_{nm\alpha},
\ee
where $\Delta_{nm\alpha}$ is defined in Theorem \ref{thm1}.
\end{thm}

\begin{thm}\label{thm6}
Assume that Assumptions \ref{ass-3-1}, \ref{ass-3-3}-\ref{ass-3-5} hold, the iterates $\{\vtheta_t\}_{t=0}^\infty$ produced by Option II in Algorithm \ref{alg2} with $\vtheta_0\in B(\hat{\vtheta}, R)$ for $t\geq0$,
$\sbr{\frac{\delta+2R^{-1}\Delta_{nm\beta}}{\rho+\lambda}}^2<\frac{\rho}{\rho+2\lambda}$, and $\alpha\leq \beta\leq \frac{1}{2}-\varepsilon$ for some $\varepsilon>0$. Then, with probability at least $1-\frac{2p(m+2)}{(1+n(m+1)\tilde{L}D)^p}$, we have
$$
\|\vtheta_{t+1}-\hat{\vtheta}\|_2
\leq \frac{\frac{\delta}{\rho+\lambda}\sqrt{\rho^2+2\lambda\rho}+\lambda}{\rho+\lambda}\|\vtheta_{t}-\hat{\vtheta}\|_2+\frac{2}{\rho+\lambda}\Delta_{nm\beta},
$$
where $\Delta_{nm\beta}$ is defined in Theorem \ref{thm2}.
\end{thm}

Theorems \ref{thm5} and \ref{thm6} present the linear convergence of Algorithm \ref{alg2} Options I and II, respectively. Obviously, the contraction factor consists of two parts: $\frac{\frac{\delta}{\rho+\lambda}\sqrt{\rho^2+2\lambda\rho}}{\rho+\lambda}$ which comes from the error of the inexact proximal update $\|\vtheta_{t+1}-\mathrm{prof}_{\lambda^{-1}(f+g)}(\vtheta_t)\|_2$, and $\frac{\lambda}{\rho+\lambda}$ which comes from the residual of proximal point $\|\mathrm{prof}_{\lambda^{-1}(f+g)}(\vtheta_t)-\hat{\vtheta}\|_2$.
Remarking similar to Theorems \ref{thm1} and \ref{thm2}, within a finite $T$-step , we have
$\|\vtheta_{T}-\hat{\vtheta}\|_2=\tilde{\mathcal{O}}_p\sbr{\frac{\alpha}{\sqrt{n}}+\frac{1}{\sqrt{nm}}+\frac{1}{n}}$ for Option I, which is a order-optimal if $n\gtrsim m$; and $\|\vtheta_{T}-\hat{\vtheta}\|_2=\mathcal{\tilde{O}}_p\sbr{\frac{\beta}{\sqrt{n}}+\frac{1}{\sqrt{nm}}}$ for Option II, which also is a order optimal. These results just need that $\{f_k\}_1^{m+1}$ are convex and smooth while the penalty $g$ is allowed to be non-smooth, for example, $\ell_1$ norm. However, Most distributed statistical learning algorithms are only designed for smooth problems, and don't consider Byzantine problems, for instance, \cite{Shamiretal2014}, \cite{WangKolarSZ2017}, \cite{JordanLeeYang2019}, and so on.

In Theorems \ref{thm1} and \ref{thm2}, we require the homogeneity assumption $\rho>\delta+2\Delta_{nm\alpha}/R$ between $f_1(\cdot)$ and $F(\cdot)$. That is, we need they must be similar enough. By the law of large numbers, the sample size of the 1st worker machine must to be large. From Theorems \ref{thm5} and \ref{thm6}, we see that such a condition is no longer needed, as long as the condition $\sbr{\frac{\delta+2R^{-1}\Delta_{nm\beta}}{\rho+\lambda}}^2<\frac{\rho}{\rho+2\lambda}$.  The condition holds definitely by
choosing sufficiently large regularity $\lambda$. Therefore, Algorithm \ref{alg2} needs a weaker homogeneous hypothesis than Algorithm \ref{alg1}. After running a finite step $T$ parallel iterations, with high probability we can obtain the error $\|\vtheta_T-\hat{\vtheta}\|_2\leq \frac{4}{\rho-\delta\sqrt{\rho^2+2\lambda\rho}/(\rho+\lambda)}\Delta_{nm\alpha}$ (Option I) or $\frac{4}{\rho-\delta\sqrt{\rho^2+2\lambda\rho}/(\rho+\lambda)}\Delta_{nm\beta}$ (Option II).
Furthermore, the choice of a large $\lambda$ can accelerate its contraction. At this time, $\delta$ has little effect on the contraction factor. This is an important aspect of Algorithm \ref{alg2} contribution.

The following corollary gives the choice of $\lambda$ that makes Algorithm \ref{alg2} converge.

\begin{cor}\label{cor4-1}
Under the assumptions of Theorem \ref{thm5} or \ref{thm6},

(i) if $\lambda>4\delta^2/\rho$, then with high probability,
$$\|\vtheta_{t+1}-\hat{\vtheta}\|_2
\leq \sbr{1-\frac{\rho}{10(\lambda+\rho)}}\|\vtheta_{t}-\hat{\vtheta}\|_2+\frac{2}{\rho+\lambda}\Lambda_{nm};$$

(ii) if $\lambda\leq C\delta^2/\rho$ for some constant $C$ and $\delta/\rho$ is sufficiently small, then with high probability,
$$\|\vtheta_{t+1}-\hat{\vtheta}\|_2
\leq O\sbr{\frac{\delta}{\rho}}\|\vtheta_{t}-\hat{\vtheta}\|_2+\frac{2}{\rho+\lambda}\Lambda_{nm};$$
where $\Lambda_{nm}=\Delta_{nm\alpha}$ for Algorithm \ref{alg1} and $\Lambda_{nm}=\Delta_{nm\beta}$ for Algorithm \ref{alg2}.
\end{cor}

From Corollary \ref{cor4-1}, we can choose $$\lambda\asymp \delta^2/\rho$$
as a default choice for Algorithm \ref{alg2} to make the algorithm converge naturally. And then we achieve the order-optimal error rate
after running a finite parallel iterations; They are $\tilde{\mathcal{O}}_p\sbr{\frac{\alpha}{\sqrt{n}}+\frac{1}{\sqrt{nm}}+\frac{1}{n}}$ for Option I, which is a order-optimal if $n\gtrsim m$, and $\mathcal{\tilde{O}}_p\sbr{\frac{\beta}{\sqrt{n}}+\frac{1}{\sqrt{nm}}}$ for Option II. From Corollary \ref{cor4-1}(ii), we see that with a regularizer $\lambda$ up to $O(\delta^2/\rho)$, the contraction fact $O\sbr{\frac{\delta}{\rho}}$ is essentially the same as the case of the unregularized problem ($\lambda=0$). It also tell us how large $\lambda$ can be chose so that the contraction factor is the same order as Algorithm \ref{alg1}. Also see Theorems
\ref{thm1} and \ref{thm2}.

\subsection{Statistical analysis for general models}
In the subsection, we consider the case of generalized linear models as in Subsection \ref{subsec3-2}. In Algorithm \ref{alg2}, $\lambda$ is a regularization parameter which is very important for adapting to the different scenarios of $n/p$. That is, by specifying the correct order of the $\lambda$, Algorithm \ref{alg2} can solve the dilemma of small local sample size $n$, while enjoying all the characteristics of Algorithm \ref{alg1} in the large-$n$ local data.

\begin{thm}\label{thm7}
Under the assumptions of Theorem \ref{thm3} or \ref{thm4}, except that with high probability $\vtheta_0\in B(\hat{\vtheta},R/2)$ for some $R>\|\hat{\vtheta}-\vtheta^*\|_2$. Let $\eta=\kappa^2p(\log p+\log n)/n$ and $\kappa=\|\Sigma\|_2/\rho$. For any $c_1, c_2>0$, there exists $c_3>0$, after $t$ parallel iterations,

(i) if $n>c_1p$ and $\alpha>c_3\rho\eta$, then the algorithms have linear convergence
$$\|\vtheta_{t}-\hat{\vtheta}\|_2
\leq \sbr{1-\frac{\rho}{10(\lambda+\rho)}}^t\|\vtheta_{0}-\hat{\vtheta}\|_2+\frac{2}{\rho(1-\eta^{1/2})}\Lambda_{nm}, \forall t\geq0;$$

(ii) if $\eta$ is sufficiently small and $\alpha\leq c_2\rho\eta$, then the algorithms have
$$\|\vtheta_{t}-\hat{\vtheta}\|_2
\leq \eta^{t/2}\|\vtheta_{0}-\hat{\vtheta}\|_2+\frac{2}{\rho(1-\eta^{1/2})}\Lambda_{nm}, \forall t\geq0;$$
where $\Lambda_{nm}=\Delta_{nm\alpha}$ for Algorithm \ref{alg1} and $\Lambda_{nm\alpha}=\Delta_{nm\beta}$ for Algorithm \ref{alg2}.
\end{thm}

For Theorem \ref{thm7}(i), we assume that $n\geq c_1p$, which is reasonable especially for many big data situations. Theorem \ref{thm7}(ii) shows that by taking $\lambda=O(\rho\eta)$, Algorithm \ref{alg2} inherits all the advantages of Algorithm \ref{alg1} in the large $n$ regime, one of them is fast linear contraction with the rate $O(\kappa\sqrt{p(\log p+\log n)/n})$. In practice, it is difficult for us to determine whether the sample size is sufficiently large, but Algorithm \ref{alg2} always guarantees the convergence via proper choice of $\lambda$. And Theorem \ref{thm7}(ii) guarantees that after running $T\geq \frac{2\log\sbr{ 2^{-1}\Lambda_{nm}^{-1}\rho(1-\eta^{1/2})\|\vtheta_0-\hat{\vtheta}\|_2}}{\log (1/\eta)}$ parallel iterations, with high probability we can obtain a solution $\tilde{\vtheta}=\vtheta_T$ with error $\|\tilde{\vtheta}-\hat{\vtheta}\|_2\leq \frac{4}{\rho(1-\eta^{1/2})}\Lambda_{nm}$. Similar to the discussion to Theorems \ref{thm3} and \ref{thm4}, we have $$\|\tilde{\vtheta}-\vtheta^*\|_2=\mathcal{O}_p(\Omega),$$
where $\mathcal{O}_p(\Omega)$ is defined in (\ref{eq3-5-1}). It means that $\|\tilde{\vtheta}-\vtheta^*\|_2=\tilde{\mathcal{O}}_p\sbr{\frac{\alpha}{\sqrt{n}}+\frac{1}{\sqrt{nm}}+\frac{1}{n}}$ for Option I, which is a order-optimal if $n\gtrsim m$, and $\|\tilde{\vtheta}-\vtheta^*\|_2=\mathcal{\tilde{O}}_p\sbr{\frac{\beta}{\sqrt{n}}+\frac{1}{\sqrt{nm}}}$ for Option II, which also is a order-optimal.

As mentioned, the distributed contraction rates depend strongly on the conditional number $\kappa$, even for generalized linear models. Here, we give another specific case of distributed linear regression on $L_2$ loss, and then obtain the strong results under some specific conditions. We define a $L_2$ loss on the $k$th worker machine as
$$f_k(\vtheta)=\frac{1}{2n}\sum_{i\in\mathcal{I}_k}(y_i-\vc x_i^T\vtheta)^2=\frac{1}{2}\vtheta^T \hat{\vc\Sigma}_k\vtheta-\hat{\vc v}_k^T\vtheta+\frac{1}{2n}\sum_{i\in\mathcal{I}_k}y_i^2,$$
where $\hat{\vc \Sigma}_k=\frac{1}{n}\sum_{i\in\mathcal{I}_k}\vc x_i\vc x_i^T$ and $\hat{\vc v}_k=\frac{1}{n}\sum_{i\in\mathcal{I}_k}\vc x_iy_i$; and
$$f(\vtheta)=\frac{1}{m+1}\sum_{k=1}^{m+1}f_k(\vtheta)=\frac{1}{2}\vtheta^T\hat{\vc\Sigma}\vtheta-\hat{\vc v}^T\vtheta+\frac{1}{2N}\sum_{i=1}^Ny_i^2,$$
where $\hat{\vc\Sigma}=\frac{1}{N}\sum_{i=1}^N\vc x_i\vc x_i^T$ and $\hat{\vc v}=\frac{1}{N}\sum_{i=1}^N\vc x_iy_i$.

For Algorithm \ref{alg2} with the designation $g(\vtheta)=0$, we have the closed form
\be\label{eq4-10}
\vtheta_{t+1}=(\hat{\vc\Sigma}_1+\lambda \vc I)^{-1}\mbr{(\hat{\vc\Sigma}_1+\lambda\vc I)\vtheta_t-\vc h(\vtheta_t)},
\ee
where $\vc h(\vtheta_t)$ is defined by (\ref{eq-alg2}).

We present the following assumptions.
\begin{ass}\label{ass-4-1}
(i) $E\vc x_i=0$ and $E(\vc x_i\vc x_i^T)=\vc\Sigma\succ0$. The minimum eigenvalue $\sigma_{\min}(\vc\Sigma)$ is bounded away from zero.

(ii) $N/\mathrm{Tr}(\vc\Sigma)\geq C>0$ and $n/\log m\geq c>0$ where $C$ and $c$ are constants.

(iii) $\{\vc \Sigma^{-1/2}\vc x_i\}_{i=1}^N$ are i.i.d. sub-Gaussian random vectors with bounded $\|\vc \Sigma^{-1/2}\vc x_i\|_{\psi_2}$.

(iv) $\{\vc \Sigma^{-1/2}\vc x_i\}_{i=1}^N$ are i.i.d. random vectors, and each component is $v$-sub-exponential.
\end{ass}

\begin{thm}\label{thm8}
Assume that there exist positive constants $C_1$ such that (1) $n>C_1p$ and $\lambda\geq0$ or (2) $\lambda\geq C_1\mathrm{Tr}(\vc\Sigma)/n$, and $n/p$ is bound away from zero. In addition,

(a) if Assumption \ref{ass-4-1}(i)-(iii) holds and the fraction $\alpha$ of Byzantine machines satisfies (\ref{eq3-0}), then with high probability,
$$
\left\|(\vtheta_{t}-\hat{\vtheta})\right\|_2\leq\sqrt{3\kappa}\sbr{1-\frac{1-\min\{1/2,C_2p/n\}}{1+C_3\lambda}}^t\left\|(\vtheta_0-\hat{\vtheta})\right\|_2\nonumber\\
+\mathcal{O}_p(N^{-1/2})+\mathcal{O}_p(\Delta_{nm\alpha});
$$

(b) If Assumption \ref{ass-4-1}(i)-(ii)(iv) holds and the $\alpha$ and the trimmed parameter $\beta$ satisfies $\alpha\leq \beta\leq \frac{1}{2}-\varepsilon$ for some $\varepsilon>0$, then with high probability,
$$
\left\|(\vtheta_{t}-\hat{\vtheta})\right\|_2\leq\sqrt{3\kappa}\sbr{1-\frac{1-\min\{1/2,C_2p/n\}}{1+C_3\lambda}}^t\left\|(\vtheta_0-\hat{\vtheta})\right\|_2\nonumber\\
+\mathcal{O}_p(N^{-1/2})+\mathcal{O}_p(\Delta_{nm\beta}),
$$
where $\kappa=\lambda_{\max}(\vc\Sigma)/\lambda_{\min}(\vc\Sigma)$, and $\Delta_{nm\alpha}$ and $\Delta_{nm\beta}$ are defined in Theorem \ref{thm1} and \ref{thm2}, respectively.
\end{thm}

Note that if $n/p$ is large enough, then
$$1-\frac{1-\min\{1/2,C_2p/n\}}{1+C_3\lambda}=\frac{C_3\lambda+C_2p/n}{1+C_3\lambda}=O(p/n)$$
by choosing $\lambda\asymp p/n)$ based on the weak requirement for regularization ($\lambda>0$, see Condition (1) in Theorem \ref{thm8}); We know that most distributed learning algorithms do not work even without the Byzantine failure if $n/p$ is not very large. But we still guarantee linear convergence with the rate
$$1-\frac{1-\min\{1/2,C_2p/n\}}{1+C_3\lambda}=1-\frac{1}{2(1+C_3C_1\mathrm{Tr}(\vc\Sigma)/n)}<1$$
by choosing $\lambda=C_1\mathrm{Tr}(\vc\Sigma)/n$ (see Condition (2) in Theorem \ref{thm8}). Further, $\mathrm{Tr}(\vc\Sigma)=O(p)$ in most scenarios, which implies $\lambda\asymp p/n$. Therefore, we choose
$$\lambda\asymp p/n$$
as a universal and adaptive choice for Algorithm \ref{alg2} regardless of the size of $n/p$. Thus, Theorem \ref{thm8} shows that no matter the size of $n/p$, proper regularization $\lambda$ always obtains linear convergence. Hence, we can handle the distributed learning problems where the amount of data per worker machine is not large enough. This situation is difficult for some algorithms (\cite{ZhangDuchiW2013}, \cite{JordanLeeYang2019}). We still achieve an order-optimal error rate up to logarithmic factors for two options of Algorithm \ref{alg2} no matter in the large sample regime or in the general regime.

Another benefit of the Algorithm \ref{alg2} is that the contraction factor in Theorem \ref{thm8} does not depend on the condition number $\kappa$ at all, and has hardly any effect on the optimal statistical rate of the Algorithm \ref{alg2}. Therefore, it helps relax the commonly used boundedness assumption on the condition number $\kappa$ in \cite{ZhangDuchiW2013}, \cite{JordanLeeYang2019} among others. Also see the remark in Theorem 3.3 of \cite{FanGuoWang2019}. But their algorithms can not handle the distributed learning of Byzantine-failure.


\section{Numerical experiments}\label{sec-NumericalExp}

\subsection{Simulation experiments}\label{sec-simulation}

In the subsection, we present several simulated examples  to illustrate the performance of our algorithms BCLS and BCLSp, which are developed in Sections \ref{sec-BCSL} and \ref{sec-BCSLp}.

First, we conduct our numerical algorithms using the distributed logistic regression. In the logistic regression model, all the observations $\{\vc x_{ij},y_{ij}\}_{i=1}^n$ for all $j\in[m]$ are generated independently from the model
\be\label{eq-sec5-model}
y_{ij}\sim \mathrm{Ber}(P_{ij}), ~\mathrm{with}~\log\frac{P_{ij}}{1-P_{ij}}=\langle\vc x_{ij},\vtheta^*\rangle,
\ee
where $\vc x_{ij}=(1,\vc u_{ij}^T)^T$ with $\vc u_{ij}\in \mathbb{R}^p$.
In our simulation, we keep the total sample size $N=18000$ and the dimension $p=100$ fixed; the covariate vector $\vc u_{ij}$ is independently generated from $N(\vc 0_p,\vc\Sigma)$ with $\vc \Sigma=\mathrm{diag}(8,4,4,2,1,\cdots,1)\in \mathbb{R}^{p\times p}$; for each replicate of the simulation, $\vtheta^*\in \mathbb{R}^{p+1}$ is a random vector with $\norm{\vtheta^*}_2=3$ whose direction is chosen uniformly at random from the sphere.

In the distributed learning, we split the whole dataset into each worker machine. According to the regimes ``large $n$", ``moderate $n$" and ``small $n$", we set the local sample size and the number of machines as $(n,m)=(900,20)$, $(450,40)$ and $(300,60)$ respectively. For our BCLS and BCLSp algorithms, we need to simulate the Byzantine failures. The $\alpha m$ worker machines are randomly chosen to be Byzantine, and one of the rest work machines as the 1st worker machine. In the experiments, we set $\alpha=20\%$. In the coordinate-wise trimmed mean, set $\beta=20\%$. For evaluating the effect of initialization on the convergence, we respectively take $\vtheta_0=\vc 0$, and $\bar{\vtheta}$ which is the local estimator based on the data of the 1st machine. They are referred to ``zero initialization" and ``good initialization", respectively.

We implement Algorithms BCLS and BCLSp based on median, trimmed mean and mean for aggregating all local gradients, which are called BCLS-md, BCLS-tr, BCLS-me, BCLSp-md, BCLSp-tr and BCLSp-me, respectively. In the first experiment, we choose $g(\cdot)=0$.
The optimizations are carried out by mini batch stochastic gradient descent in the above algorithms. The estimation error $\|\vc\theta_t-\vc\theta^*\|_2$ is used to measure the performance of these different algorithms based on the 50 simulation replications.

\begin{figure}[ht]
\centering
\includegraphics[scale=0.45]{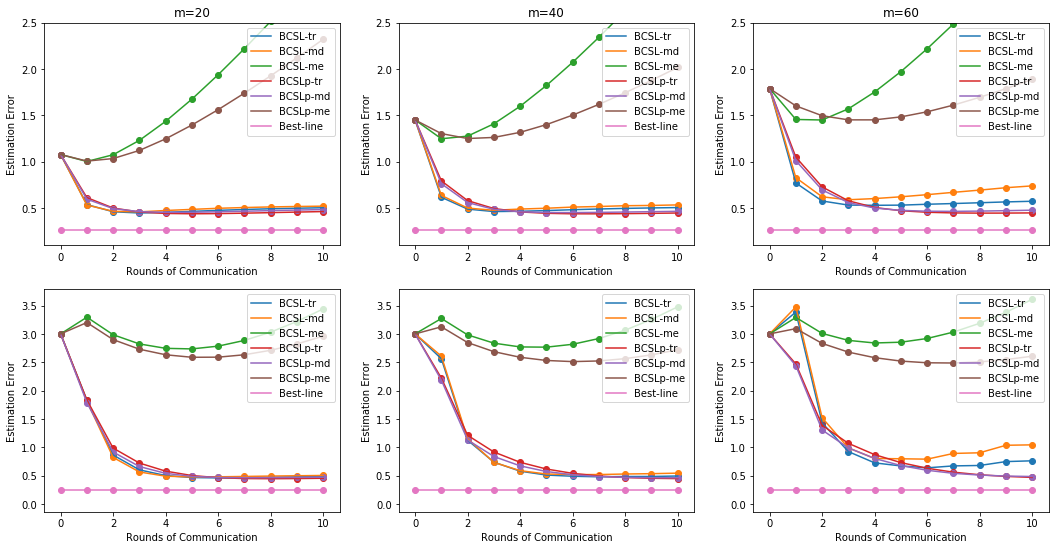}
\caption{Impacts of local sample and initialization on convergence of algorithms for the fixed $p=100$. The $x$-axis is the number of iterations or the rounds of communications, and $y$-axis is the estimation error $\|\vc\theta_t-\vc\theta^*\|_2$. The ``Best-line" shows the error of the minimizer of the overall loss function. Top panel uses $\boldsymbol{\bar{\theta}}$ as the initial value (``good initialization") and bottom uses $\boldsymbol 0$ as the initial value (``zero initialization").}
\label{figure1}
\end{figure}

Figure \ref{figure1} shows how the estimation errors evolve with iterations for the case of $p=100$ fixed. We find that our algorithms (BCLS-tr, BCLS-md, BCLSp-tr and BCLSp-md) converge reapidly in all scenarios, but the mean methods (BCLS-me and BCLSp-me) do not converge at all. Our algorithms almost converge after 2 iterations and do not require good initialization, which are in line with our theoretical results. This implies that our algorithms are robust against Byzantine failures, but the mean methods can't tolerate such failures. In addition, our proposed Algorithm BCLSp is more robust and stable than Algorithm BCLS, especially for large $m$. Note that large $m$ implies the more Byzantine worker machines. The embedding of the proximal technique in Algorithm BCLSp adds strict convex quadratic regularization, which leads to better performance.

Next, we use the distributed logistic regression model (\ref{eq-sec5-model}) with sparsity. The experiment is used to validate the efficiency of our algorithms in the presence of a nonsmooth penalty. In the simulation, we still set the total sample size $N=18000$, but the dimensionality of $\vtheta^*$ is fixed to $p=1000$; the covariate vector $\vc u_{ij}$ is i.i.d. $N(\vc 0, \vc I_p)$ and  $\vtheta^*=(\vc v_{10}^T,\vc 0_{991}^T)^T\in \mathbb{R}^{1001}$ with $\vc v_{10}\sim N(\vc 0_{10},\vc I_{10})$. The $\ell_2$-norm of $\vtheta^*$ is constrained to 3. We choose the penalty function $g(\vtheta)=\gamma\|\boldsymbol{\theta}\|_1$  with $\gamma=0.2\sqrt{{\frac{\log{p}}{N}}}$ so that the nonzeros of $\vtheta^*$ can be recovers accurately by the regularized maximum likelihood estimation over the whole dataset. As in the first experiment, we set $(n,m)=(900,20)$, $(450,40)$ and $(300,60)$, $\alpha=20\%$, $\beta=20\%$, $\vtheta_0=\vc 0$ (``zero initialization") and $\bar{\vtheta}$ (``good initialization"). The $\bar{\vtheta}$ is the local estimator based on the dataset of the 1st machine. In the Algorithm BCLSp, the penalty parameters are selected appropriately for the cases of ``good initialization" and ``zero initialization", respectively. The optimizations are carried out by mini batch stochastic sub-gradient descent in the algorithms. All the results are average values of 20 independent runs.

\begin{figure}[ht]
\centering
\includegraphics[scale=0.45]{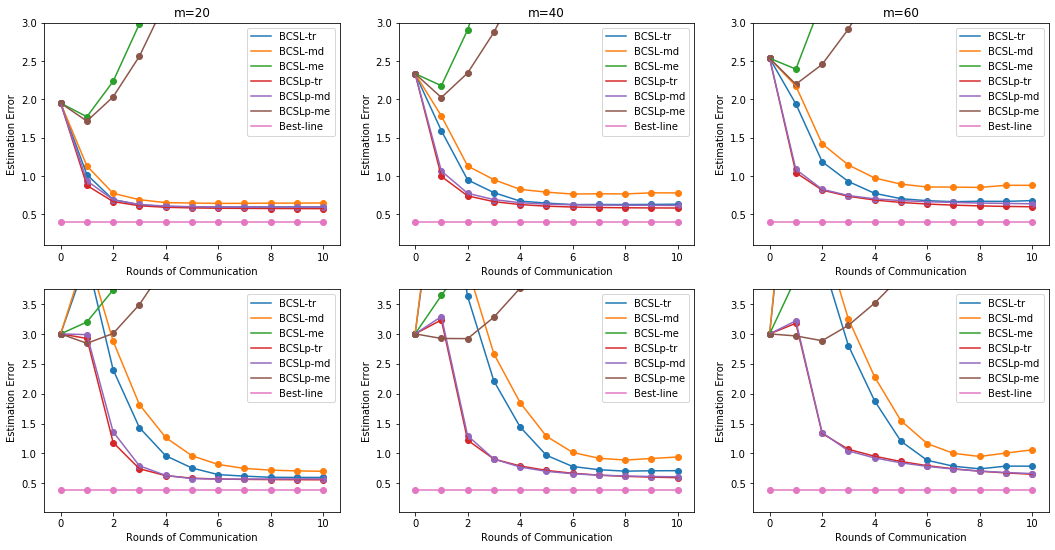}
\caption{Performance of the nonsmooth regularized algorithms for the distributed sparse logistic regression. The $x$-axis is the number of iterations or the rounds of communications, and $y$-axis is the estimation error $\|\vc\theta_t-\vc\theta^*\|_2$. The ``Best-line" shows the error of the minimizer of the overall loss function. Top panel uses $\boldsymbol{\bar{\theta}}$ as the initial value (``good initialization") and bottom uses $\boldsymbol 0$ as the initial value (``zero initialization").}
\label{figure2}
\end{figure}

Figure \ref{figure2} presents the performance of our algorithms and the mean aggregate method (BCLS-me and BCLSp-me). With proper regularization, our algorithms still work well whether the initial value is ``good" or ``bad". The mean aggregate methods (BCLS-me and BCLSp-me) fail to converge. For this nonsmooth problem, Algorithms BCLSp-tr and BCLSp-md are more robust than Algorithms BCLS-tr and BCLS-md, and start to converge after just 2 rounds of communication, specially for the case of ``good initialization".

Here, we summarize the above simulations to highlight several outstanding advantages of our algorithms.

(1) Our proposed Byzantine-Robust CSL distributed learning algorithm (BCLS-md, BCLSp-tr) and Byzantine-Robust CSL-proximal distributed learning algorithm (BCLSp-md and BCLSp-tr) can indeed defend against Byzantine failures.

(2) Our proposed algorithms converge rapidly, usually with several rounds of communication, and do not require good initialization; these are consistent with our statistical theory.

(3) The Algorithms BCLSp-md and BCLSp-tr are more robust than Algorithms BCLS-md and BCLS-tr, due to add strict convex quadratic regularization by embedding proximal technique into Algorithm BCLSp.

\subsection{Real data}
In the subsection, we further assess the performance of our proposed algorithms by a real data example. We choose Spambase dataset from the UC Irvine Machine Learning Repository \citep{DuaGraff2017}. The collection of Span e-mails in Spambase dataset come from their postmaster and individuals who had filed spam, and the collection of non-spam e-mails came from filed work and personal e-mails. Number of instances (total sample) is 4600, and Number of Attributes (features) is 57 based on their word frequencies and other characteristics. The goal is to use distributed logistic regression to construct a personalized spam filter that distinguishes spam emails from normal ones. In the experiment, randomly selecting 1000 instances as the testing set and the rest of 3600 instances as the training set; split the training set to each worker machine according to $(n,m)=(180,20)$ (``small $m$"), $(120,30)$ (``moderate $m$") and $(90,40)$ (``large $m$"), respectively; set $\alpha=20\%$ for the fraction of Byzantine worker machines and $\beta=20\%$ for the BCLS-tr and BCLSp-tr algorithms. We use classification errors on the test set as the evaluation criteria.

Figure 3 shows the average performance of the 6 algorithms mentioned in Subsection \ref{sec-simulation}. We find that the testing errors of our algorithms are very low. It implies our algorithms can accurately filter spam and non spam, even for more Byzantine worker machines (``large $m$"). But the filters based on Algorithms BCLS-me and BCLSp-me fail. These results are consistent with the ones of simulation experiments. Totally, the experiments on the real data also support our theoretical findings.

\begin{figure}[ht]
\centering
\includegraphics[scale=0.45]{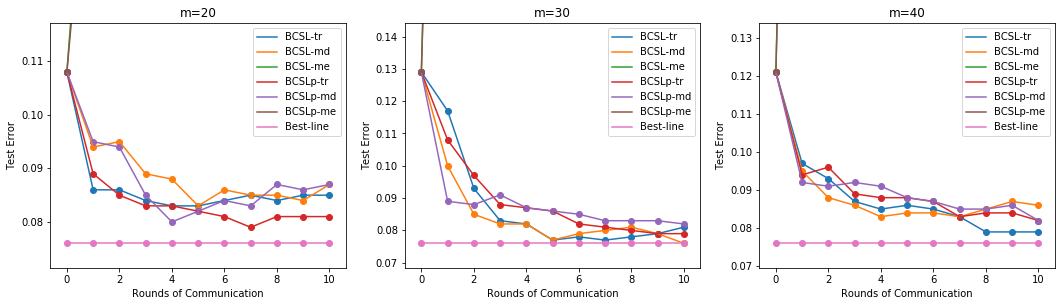}
\caption{Experiments on Spambase dataset. The $x$-axis is the number of iterations or the rounds of communications, and $y$-axis is testing error (prediction accuracy).
The ``Best-line" show the error of the filter based on all of the training dataset.}
\label{}
\end{figure}


\vskip 0.5cm
{\noindent\bf\large Acknowledgments}

This work is partially supported by Chinese National Social Science Fund (No. 19BTJ034).

\bibliography{ML-xczhou}

\newpage
\def\theequation{A.\arabic{equation}}
\setcounter{equation}{0}
\def\thesubsection{A.\arabic{subsection}}
\setcounter{subsection}{0}

\section*{Appendix A}

\subsection{Proof of Theorems \ref{thm1} and \ref{thm2}}
\noindent{\bf Proof of Theorem \ref{thm1}:}
Define
$$\varphi(\vc\xi)=\mathrm{argmin}_\vtheta\left\{f_1(\vtheta)-\langle\vc \nabla f_1(\vc\xi)-\vc h(\vc\xi),\vtheta\rangle+g(\vtheta)\right\}.$$
Obviously, $\vtheta_{t+1}=\varphi(\vtheta_t)$. By Theorem 8 in \cite{YinChenRB2018} and the law of large numbers, for $\vc\xi\in \vc\Theta$, we have
\bes
&&f_1(\vtheta)-\langle\vc \nabla f_1(\vc\xi)-\vc h(\vc\xi),\vtheta\rangle+g(\vtheta)\nonumber\\
&=&\bbr{f_1(\vtheta)-\langle\vc \nabla f_1(\vc\xi)-\nabla f(\vc\xi),\vtheta\rangle+g(\vtheta)}+[\vc h(\vc\xi)-\nabla F(\vc\xi)]+[\nabla F(\vc\xi)-\nabla f(\vc\xi)]\nonumber\\
&=&\bbr{f_1(\vtheta)-\langle\vc \nabla f_1(\vc\xi)-\nabla f(\vc\xi),\vtheta\rangle+g(\vtheta)}+o_p(1).
\ees
Therefore,  $\varphi(\hat{\vtheta})=\hat{\vtheta}+o_p(1)$, that is, $\hat{\vtheta}$ is a asymptotic fixed point of $\varphi(\cdot)$ . For the fixed $\vtheta\in B(\hat{\vtheta},R)$, by the first order condition of $\varphi(\vtheta)$, we have that
\be\label{eq3-1}
\nabla f_1(\vtheta)-\vc h(\vtheta)\in \partial\bbr{f_1(\varphi(\vtheta))+g(\varphi(\vtheta))}.
\ee
Further,
\be\label{eq3-2}
\nabla f_1(\hat{\vtheta})-\vc h(\hat{\vtheta})\in \partial\bbr{f_1(\hat{\vtheta})+g(\hat{\vtheta})},
\ee
by using the fact $\varphi(\hat{\vtheta})=\hat{\vtheta}$.

By the Taylor expansion, Assumption \ref{ass-3-4} and Theorem 8 in \cite{YinChenRB2018}, with probability at least $1-\frac{4p}{(1+n(m+1)\tilde{L}D)^p}$, we have
\bes\label{eq3-3}
&&\|[\nabla f_1(\vtheta)- h(\vtheta)]-[\nabla f_1(\hat{\vtheta})- h(\hat{\vtheta})]\|_2\nonumber\\
&\leq&\|[\nabla f_1(\vtheta)-\nabla F(\vtheta)]-[\nabla f_1(\hat{\vtheta})-\nabla F(\hat{\vtheta})]\|_2+\|[\nabla F(\vtheta)-h(\vtheta)]-[\nabla F(\hat{\vtheta})-h(\hat{\vtheta})]\|_2\nonumber\\
&\leq&\|[\nabla f_1(\vtheta)-\nabla f_1(\hat{\vtheta})]-[\nabla F(\vtheta)-\nabla F(\hat{\vtheta})]\|_2+2\Delta_{nm\alpha}\nonumber\\
&=&\left\|\int_0^1\sbr{\nabla^2f_1[(1-t)\hat{\vtheta}+t\vtheta]-\nabla^2F[(1-t)\hat{\vtheta}+t\vtheta]}(\vtheta-\hat{\vtheta})dt\right\|_2+2\Delta_{nm\alpha}\nonumber\\
&\leq& \sup_{\zeta\in B(\hat{\vtheta},R)}\|\nabla^2f_1(\zeta)-\nabla^2F(\zeta)\|_2\|\vtheta-\hat{\vtheta}\|_2+2\Delta_{nm\alpha}\nonumber\\
&\leq& \delta\|\vtheta-\hat{\vtheta}\|_2+2\Delta_{nm\alpha}\leq \delta R+2\Delta_{nm\alpha}=(\delta+2\Delta_{nm\alpha}/R)R<\rho R.
\ees

Next, we will show that under (\ref{eq3-1})-(\ref{eq3-3}), we have
\be\label{eq3-4}
\|\varphi(\vtheta)-\hat{\vtheta}\|_2\leq \left\|[\nabla f_1(\vtheta)- h(\vtheta)]-[\nabla f_1(\hat{\vtheta})- h(\hat{\vtheta})]\right\|_2/\rho.
\ee

If we know $\|\varphi(\vtheta)-\hat{\vtheta}\|_2\leq R$ in advance, then
\bes
\rho\|\varphi(\vtheta)-\hat{\vtheta}\|_2^2&\leq& \left\langle [\nabla f_1(\vtheta)- h(\vtheta)]-[\nabla f_1(\hat{\vtheta})- h(\hat{\vtheta})], \varphi(\theta)-\hat{\vtheta} \right\rangle\nonumber\\
&\leq&\|[\nabla f_1(\vtheta)- h(\vtheta)]-[\nabla f_1(\hat{\vtheta})- h(\hat{\vtheta})]\|_2\|\varphi(\theta)-\hat{\vtheta}\|_2.\nonumber
\ees
Here the 1st inequality follows from the $\rho$-strong convex in $B(\hat{\vtheta},R)$ of $f_1+g$ and (\ref{eq3-1})-(\ref{eq3-2}); the 2nd step uses the Cauchy-Schwarz inequality. Then, we obtain the desired result (\ref{eq3-4}). Suppose on the contrary that $\|\varphi(\vtheta)-\hat{\vtheta}\|_2> R$. We use reduction to absurdity. Define $\varphi(\bar{\vtheta})=\hat{\vtheta}+R(\varphi(\vtheta)-\hat{\vtheta})/\|\varphi(\vtheta)-\hat{\vtheta}\|_2$. Obviously, $\|\varphi(\bar{\vtheta})-\hat{\vtheta}\|_2=R$. By the strong convexity of $f_1+g$ in $B(\hat{\vtheta},R)$ and (\ref{eq3-1})-(\ref{eq3-2}) again,
\be\label{eq3-5}
\left\langle [f_1(\bar{\vtheta})-h(\bar{\vtheta})]-[f_1(\hat{\vtheta})-h(\hat{\vtheta})], \varphi(\bar{\vtheta})-\hat{\vtheta} \right\rangle\geq \rho\|\varphi(\bar{\vtheta})-\hat{\vtheta}\|_2^2;
\ee
Notice that
$$\varphi(\bar{\vtheta})-\hat{\vtheta}=\frac{R}{\|\varphi(\vtheta)-\hat{\vtheta}\|_2-R}(\varphi(\vtheta)-\varphi(\bar{\vtheta})).$$
Thus, by the convexity of $f_1+g$ and (\ref{eq3-1}), we always have
\bes\label{eq3-6}
&&\left\langle[f_1({\vtheta})-h({\vtheta})]-[f_1(\bar{\vtheta})-h(\bar{\vtheta})],\varphi(\bar{\vtheta})-\hat{\vtheta}\right\rangle\nonumber\\
&=&\frac{R}{\|\varphi(\vtheta)-\hat{\vtheta}\|_2-R}\left\langle[f_1({\vtheta})-h({\vtheta})]-[f_1(\bar{\vtheta})-h(\bar{\vtheta})],\varphi(\vtheta)-\varphi(\bar{\vtheta})\right\rangle\geq 0,
\ees
for any $\bar{\vtheta}\in B(\hat{\vtheta},R)$.
Summing up the (\ref{eq3-5}), and by (\ref{eq3-6}) and Cauchy-Schwarz inequality, one gets
\bes
\rho\|\varphi(\bar{\vtheta})-\hat{\vtheta}\|_2^2&\leq&\left\langle[f_1({\vtheta})-h({\vtheta})]-[f_1(\hat{\vtheta})-h(\hat{\vtheta})],\varphi(\bar{\vtheta})-\hat{\vtheta}\right\rangle\nonumber\\
&\leq&\|[f_1({\vtheta})-h({\vtheta})]-[f_1(\hat{\vtheta})-h(\hat{\vtheta})]\|_2\|\varphi(\bar{\vtheta})-\hat{\vtheta}\|_2.\nonumber
\ees
Thus, $\|[f_1({\vtheta})-h({\vtheta})]-[f_1(\hat{\vtheta})-h(\hat{\vtheta})]\|_2\geq \rho\|\varphi(\bar{\vtheta})-\hat{\vtheta}\|_2=\rho R$, which contradicts (\ref{eq3-3}). Therefore, we must have only $\|\varphi(\vtheta)-\hat{\vtheta}\|_2\leq R$. And then (\ref{eq3-4}) holds. Together with (\ref{eq3-3}), with probability at least $1-\frac{4p}{(1+n(m+1)\tilde{L}D)^p}$, we have
$$
\|\varphi(\vtheta)-\hat{\vtheta}\|_2\leq\frac{\delta}{\rho}\|\vtheta-\hat{\vtheta}\|_2+\frac{2}{\rho}\Delta_{nm\alpha}.
$$
Take $\vtheta=\vtheta_t$, then $\varphi(\vtheta_t)=\vtheta_{t+1}$. We complete the proof of Theorem \ref{thm1}.

\noindent{\bf Proof of Theorem \ref{thm2}:} The proof is essentially the same as the proof of Theorem \ref{thm1}, except that  the analysis of coordinate-wise trimmed mean of means estimator of the population gradients, which can be found in \cite{YinChenRB2018}, is different one of coordinate-wise median.

\subsection{Proof of Theorems \ref{thm3} and \ref{thm4}}

Theorems \ref{thm3} and \ref{thm4} are the special cases of Theorem \ref{thm7} (ii) by taking $\lambda=0$. See subsection A. for the proofs of Theorem \ref{thm7}.

\subsection{Proofs of Theorems \ref{thm5}, \ref{thm6} and Corollary \ref{cor4-1}}

\noindent{\bf Proof of Theorem \ref{thm5}:} Under the conditions of Theorem \ref{thm5}, if $0<\|\vtheta_t-\hat{\vtheta}\|_2<R/2$, then (\ref{eq4-2}) holds. Then Theorem \ref{thm5} can directly follow from the result and induction. Below we prove the result.

Recall that $\hat{\vtheta}=\mathrm{argmin}_\vtheta(f(\vtheta)+g(\vtheta))$. Denote $\vtheta_t^+=\mathrm{prox}_{\lambda^{-1}(f+g)}(\vtheta_t)$. By the triangle inequality,
\be\label{eq4-3}
\|\vtheta_{t+1}-\hat{\vtheta}\|_2\leq \|\vtheta_{t+1}-\vtheta_t^+\|_2+\|\vtheta_t^+-\hat{\vtheta}\|_2.
\ee
For the first term on the right of (\ref{eq4-3}), $\|\vtheta_{t+1}-\vtheta_t^+\|_2$, we can obtain its contracting optimization errors by Theorem \ref{thm1}, taking $g(\vtheta)$ in Algorithm \ref{alg1} as $\tilde{g}(\vtheta)=g(\vtheta)+\frac{\lambda}{2}\|\vtheta-\vtheta_t\|_2^2$.
Thus, $\vtheta_t^+=\mathrm{argmin}_\vtheta\bbr{f(\vtheta)+\tilde{g}(\vtheta)}$. Together with (\ref{eq-alg2}), $\vtheta_{t+1}$ is regarded as the first iterate of Algorithm \ref{alg1} initialized at $\vtheta_t$ for obtaining $\vtheta_t^+$. Here, $\vtheta_t^+$ is similar to $\hat{\vtheta}$ in Algorithm \ref{alg1}. Contrasting the assumptions of Theorem \ref{thm1}, we still need the following assertions:

(i) $\vtheta_t\in B(\vtheta_t^+,R/2)$;

(ii) $f+\tilde{g}$ is $\rho+\lambda$-strongly convex in $B(\vtheta_t^+, R/2)$;

(iii) $\|\nabla^2 f_1(\vtheta)-\nabla^2 F(\vtheta)\|_2\leq \delta$, and $\vtheta\in B(\vtheta_t^+, R/2)$.

Let $\tilde{f}=f+g$. Notice that
\be\label{eq4-4}
\hat{\vtheta}=\mathrm{prox}_{\lambda^{-1}\tilde{f}}(\hat{\vtheta}).
\ee
By the well-known ``firm non-expansiveness" property of the proximal operation \citep{ParikhBoyd2014}, we have
\be\label{eq4-5}
\left\|\mathrm{prox}_{\lambda^{-1}\tilde{f}}(\vtheta_t)-\mathrm{prox}_{\lambda^{-1}\tilde{f}}(\hat{\vtheta})\right\|_2^2
\leq\left\langle\vtheta_t-\hat{\vtheta},\mathrm{prox}_{\lambda^{-1}\tilde{f}}(\vtheta_t)-\mathrm{prox}_{\lambda^{-1}\tilde{f}}(\hat{\vtheta})\right\rangle.
\ee
From (\ref{eq4-4})-(\ref{eq4-5}), we have $
\|\vtheta_t^+-\hat{\vtheta}\|_2=\|\mathrm{prox}_{\lambda^{-1}\tilde{f}}(\vtheta_t)-\hat{\vtheta}\|_2
\leq\|\vtheta_t-\hat{\vtheta}\|_2
$. So, the condition $\|\vtheta_t-\hat{\vtheta}\|_2<R/2$ implies $B(\vtheta_t^+,R/2)\subset B(\hat{\vtheta},R)$. Assumptions \ref{ass-3-3} and \ref{ass-3-4} imply (ii) and (iii) hold, respectively. In the other hand,
\bes
&&\left\|\mathrm{prox}_{\lambda^{-1}\tilde{f}}(\vtheta_t)-\vtheta_t\right\|_2^2=
\left\|[\mathrm{prox}_{\lambda^{-1}\tilde{f}}(\vtheta_t)-\hat{\vtheta}]-[\vtheta_t-\hat{\vtheta}]\right\|_2^2\nonumber\\
&=&\left\|\mathrm{prox}_{\lambda^{-1}\tilde{f}}(\vtheta_t)-\hat{\vtheta}\right\|_2^2+\|\vtheta_t-\hat{\vtheta}\|_2^2
-2\left\langle\mathrm{prox}_{\lambda^{-1}\tilde{f}}(\vtheta_t)-\hat{\vtheta},\vtheta_t-\hat{\vtheta}\right\rangle\nonumber\\
&\leq&\left\|\mathrm{prox}_{\lambda^{-1}\tilde{f}}(\vtheta_t)-\hat{\vtheta}\right\|_2^2+\|\vtheta_t-\hat{\vtheta}\|_2^2
-2\left\|\mathrm{prox}_{\lambda^{-1}\tilde{f}}(\vtheta_t)-\hat{\vtheta}\right\|_2^2\nonumber\\
&=&\|\vtheta_t-\hat{\vtheta}\|_2^2
-\left\|\mathrm{prox}_{\lambda^{-1}\tilde{f}}(\vtheta_t)-\hat{\vtheta}\right\|_2^2,\nonumber
\ees
that is,
$$
\|\vtheta_t^+-\vtheta_t\|_2^2\leq\|\vtheta_t-\hat{\vtheta}\|_2^2-\|\vtheta_t^+-\hat{\vtheta}\|_2^2.
$$
Further,
\be\label{eq4-6}
\|\vtheta_t^+-\vtheta_t\|_2\leq\|\vtheta_t-\hat{\vtheta}\|_2\sqrt{1-\|\vtheta_t^+-\hat{\vtheta}\|_2^2/\|\vtheta_t-\hat{\vtheta}\|_2^2},
\ee
which leads to $\|\vtheta_t^+-\vtheta_t\|_2\leq\|\vtheta_t-\hat{\vtheta}\|_2<R/2$.
Therefore, (i) holds.

Based the above assertions and $\rho+\lambda>\delta+2\Delta_{nm\alpha}/R$ by $\sbr{\frac{\delta+2R^{-1}\Delta_{nm\alpha}}{\rho+\lambda}}^2<\frac{\rho}{\rho+2\lambda}$, by Theorem \ref{thm1}, we have
\be\label{eq4-7}
\|\vtheta_{t+1}-\vtheta_t^+\|_2\leq \frac{\delta}{\rho+\lambda}\|\vtheta_t-\vtheta_t^+\|_2+\frac{2}{\rho+\lambda}\Delta_{nm\alpha}.
\ee
From (\ref{eq4-3}), (\ref{eq4-6}) and (\ref{eq4-7}), we have
\bes\label{eq4-8}
\|\vtheta_{t+1}-\hat{\vtheta}\|_2&\leq& \frac{\delta}{\rho+\lambda}\|\vtheta_t-\vtheta_t^+\|_2+\frac{2}{\rho+\lambda}\Delta_{nm\alpha}+\|\vtheta_t^+-\hat{\vtheta}\|_2\nonumber\\
&\leq&\|\vtheta_t-\hat{\vtheta}\|_2\sbr{\frac{\delta}{\rho+\lambda}\sqrt{1-\frac{\|\vtheta_t^+-\hat{\vtheta}\|_2^2}{\|\vtheta_t-\hat{\vtheta}\|_2^2}}
+\frac{\|\vtheta_t^+-\hat{\vtheta}\|_2}{\|\vtheta_t-\hat{\vtheta}\|_2}}+\frac{2}{\rho+\lambda}\Delta_{nm\alpha}\nonumber\\
&=&\|\vtheta_t-\hat{\vtheta}\|_2\kappa\sbr{\frac{\|\vtheta_t^+-\hat{\vtheta}\|_2}{\|\vtheta_t-\hat{\vtheta}\|_2}}+\frac{2}{\rho+\lambda}\Delta_{nm\alpha},
\ees
where $\kappa(u)=\frac{\delta}{\rho+\lambda}\sqrt{1-u^2}+u$.

Here, we will prove
\be\label{eq4-9}
\frac{\|\vtheta_t^+-\hat{\vtheta}\|_2}{\|\vtheta_t-\hat{\vtheta}\|_2}\leq \frac{\lambda}{\rho+\lambda}<1,
\ee
which makes (\ref{eq4-8}) valid. Indeed, on the hand, $\|\vtheta_t^+-\hat{\vtheta}\|_2
\leq\|\vtheta_t-\hat{\vtheta}\|_2$; on the other hand, $\hat{\vtheta}=\mathrm{argmin}_\vtheta\tilde{f}(\vtheta)$ and $\vtheta_t^+=\mathrm{argmin}_\vtheta\sbr{\tilde{f}(\vtheta)+\lambda/2\|\vtheta-\vtheta_t\|_2^2}$ imply that $\vc 0\in \partial\tilde{f}(\hat{\vtheta})$ and $-\lambda(\vtheta_t^+-\vtheta_t)\in\partial\tilde{f}(\vtheta_t^+)$.
Because $\tilde{f}$ is $\rho$-strongly convex in $B(\hat{\vtheta},R/2)$ and the basic properties of strong convex functions \citep{Nesterov2004}, we have
\bes
\rho\|\vtheta_t^+-\hat{\vtheta}\|_2^2&\leq& \langle-\lambda(\vtheta_t^+-\vtheta_t)-\vc 0,\vtheta_t^+-\hat{\vtheta}\rangle\nonumber\\
&=&-\lambda\|\vtheta_t^+-\hat{\vtheta}\|_2^2-\lambda\langle\hat{\vtheta}-\vtheta_t,\vtheta_t^+-\hat{\vtheta}\rangle\nonumber\\
&\leq&-\lambda\|\vtheta_t^+-\hat{\vtheta}\|_2^2+\lambda\|\hat{\vtheta}-\vtheta_t\|_2\|\vtheta_t^+-\hat{\vtheta}\|_2.\nonumber
\ees
Thus, we get (\ref{eq4-9}).

For $\kappa(u)$ in (\ref{eq4-9}), $0\leq u<1$ and $\kappa(u)$ is an increasing function on $[0, 1/\sqrt{1+[\delta/(\rho+\lambda)]^2}]$.
Notice that $\sbr{\frac{\delta+2R^{-1}\Delta_{nm\alpha}}{\rho+\lambda}}^2<\frac{\rho}{\rho+2\lambda}$ implies $\sbr{\frac{\delta}{\rho+\lambda}}^2<\frac{\rho}{\rho+2\lambda}$. Therefore,
$$\frac{1}{\sqrt{1+[\delta/(\rho+\lambda)]^2}}>\frac{\sqrt{\lambda+\rho/2}}{\sqrt{\rho+\lambda}}\geq \frac{\lambda+\rho/2}{\rho+\lambda}\geq \frac{\lambda}{\rho+\lambda}.$$
Then, by (\ref{eq4-9}) and $\sbr{\frac{\delta}{\rho+\lambda}}^2<\frac{\rho}{\rho+2\lambda}$,
\bes
\kappa\sbr{\frac{\|\vtheta_t^+-\hat{\vtheta}\|_2}{\|\vtheta_t-\hat{\vtheta}\|_2}}&\leq&\kappa\sbr{\frac{\lambda}{\rho+\lambda}}
=\frac{\delta}{\rho+\lambda}\sqrt{1-\sbr{\frac{\lambda}{\rho+\lambda}}^2}+\frac{\lambda}{\rho+\lambda}\nonumber\\
&=&\frac{\frac{\delta}{\rho+\lambda}\sqrt{\rho^2+2\rho\lambda}+\lambda}{\rho+\lambda}\nonumber\\
&=&\mbr{\sqrt{\sbr{\frac{\delta}{\rho+\lambda}}^2\rho(\rho+2\lambda)}+\lambda}/(\rho+\lambda)<1.\nonumber
\ees
Combining with (\ref{eq4-8}), we complete the proof of Theorem \ref{thm5}.

\noindent{\bf Proof of Theorem \ref{thm6}:} The proof is essentially the same as the proof of Theorem \ref{thm5}, but invoke Theorem \ref{thm3} to bound $\|\vtheta_{t+1}-\vtheta_t^+\|_2$ in (\ref{eq4-3}).

\noindent{\bf Proof of Corollary \ref{cor4-1}:} The proof is similar to the ones of Corollaries 3.1 and 3.2 in \cite{FanGuoWang2019}. Just a few algebraic tricks are needed. Here, we omit the details.

\subsection{Proofs of Theorems \ref{thm7} and \ref{thm8}}

\noindent{\bf Proof of Theorem \ref{thm7}:} The proof is implied by combining proof of Corollary \ref{cor4-1} with Lemma A.5 in \cite{FanGuoWang2019}, which provides the order of Hessian difference on the 1st worker machine in the GLM. Further, it presents a contraction rate and a choice of $\lambda$.

\noindent{\bf Proof of Theorem \ref{thm8}:} We only prove (a). Proof of (b) is similar to (a) by applying Bernstein's inequality for sub-exponential random variables. Let $\vc\Xi=(\hat{\vc\Sigma}+\lambda \vc I)^{-1/2}\sbr{\hat{\vc\Sigma}_1-\hat{\vc\Sigma}}\sbr{\hat{\vc\Sigma}+\lambda \vc I}^{-1/2}$.

First, we have the following basic facts:

(i) $P\{\vc\Xi\leq 1/2\}\geq 1-2e^{-n/C}$ for some constant $C$, which is determined by $\|\vc x_i\|_{\psi_2}$, under the condition (1) $n>C_1p$ and $\lambda\geq0$ or (2) $\lambda\geq C_1\mathrm{Tr}(\vc\Sigma)/n$.

(ii) $P\left\{\left\|\vc\Sigma^{-1/2}(\hat{\vc\Sigma}-\vc\Sigma)\vc\Sigma^{-1/2}\right\|_2\leq 1/2\right\}\geq 1-2e^{N/C}$ for some constant $C$ in (i).

(iii) $P\{\vc\Xi\leq C_2\sqrt{p/n}\}\geq 1-2e^{-C_1p}$ for some constants $C_1$ and $C_2$.

By the fact (i), we can appropriately choose $\lambda$, such that with high probability $\vc\Xi\leq 1/2$. Then

(iv) $\left\|\vc I-\hat{\vc\Sigma}^{1/2}(\hat{\vc\Sigma}_1+\lambda \vc I)^{-1}\hat{\vc\Sigma}^{1/2}\right\|_2\leq \frac{2\vc\Xi^2+\lambda/\lambda_{\min}(\hat{\vc\Sigma})}{1+\lambda/\lambda_{\min}(\hat{\vc\Sigma})}$, for any $t\geq 0$.\\
These facts can found in Lemmas A.6-A.8 of \cite{FanGuoWang2019}.

Now we give the main proof. From (\ref{eq4-10}), the law of large numbers and Theorem 8 in \cite{YinChenRB2018}, we have
\bes
\vtheta_{t+1}&=&(\hat{\vc\Sigma}_1+\lambda \vc I)^{-1}\mbr{(\hat{\vc\Sigma}_1+\lambda\vc I)\vtheta_t-\nabla f(\vtheta)}\nonumber\\
&&+(\hat{\vc\Sigma}_1+\lambda \vc I)^{-1}\mbr{\nabla f(\vtheta)-\nabla F(\vtheta)}+(\hat{\vc\Sigma}_1+\lambda \vc I)^{-1}\mbr{\nabla F(\vtheta)-\vc h(\vtheta_t)}\nonumber\\
&=&[\vc I-(\hat{\vc\Sigma}_1+\lambda \vc I)^{-1}\hat{\vc\Sigma}]\vc\vtheta_t+(\hat{\vc\Sigma}_1+\lambda \vc I)^{-1}\hat{\vc v}
+\mathcal{O}_p(N^{-1/2})+\mathcal{O}_p(\Delta_{nm\alpha}).\nonumber
\ees
Further, one gets
\bes
&&\hat{\vc\Sigma}^{1/2}(\vtheta_{t+1}-\hat{\vtheta})=\hat{\vc\Sigma}^{1/2}(\vtheta_{t+1}-\hat{\vc\Sigma}^{-1}\hat{\vc v})\nonumber\\
&=&\hat{\vc\Sigma}^{1/2}[\vc I-(\hat{\vc\Sigma}_1+\lambda \vc I)^{-1}\hat{\vc\Sigma}]\vc\vtheta_t+\hat{\vc\Sigma}^{1/2}(\hat{\vc\Sigma}_1+\lambda \vc I)^{-1}\hat{\vc v}-\hat{\vc\Sigma}^{-1/2}\hat{\vc v}\nonumber\\
&&+\mathcal{O}_p(N^{-1/2})+\mathcal{O}_p(\Delta_{nm\alpha})\nonumber\\
&=&[\vc I-\hat{\vc\Sigma}^{1/2}(\hat{\vc\Sigma}_1+\lambda \vc I)^{-1}\hat{\vc\Sigma}^{1/2}]\hat{\vc\Sigma}^{1/2}(\vtheta_{t}-\hat{\vtheta})+\mathcal{O}_p(N^{-1/2})+\mathcal{O}_p(\Delta_{nm\alpha}).\nonumber
\ees
Together with (iv), we have
\be\label{eq4-11}
\left\|\hat{\vc\Sigma}^{1/2}(\vtheta_{t+1}-\hat{\vtheta})\right\|_2\leq \frac{2\vc\Xi^2+\lambda/\lambda_{\min}(\hat{\vc\Sigma})}{1+\lambda/\lambda_{\min}(\hat{\vc\Sigma})}\left\|\hat{\vc\Sigma}^{1/2}(\vtheta_t-\hat{\vtheta})\right\|_2+\mathcal{O}_p(N^{-1/2})+\mathcal{O}_p(\Delta_{nm\alpha}).
\ee
From (i)-(iii), we have $\vc\Xi\leq 1/2$, $\vc\Xi\leq C_2\sqrt{p/n}$ and $\lambda_{\min}(\hat{\vc\Sigma})\geq C_3^{-1}>0$ simultaneously with high probability, where $C_2$ and $C_3$ are some positive constants. So, with high probability, we have
\be\label{eq4-12}
\frac{2\vc\Xi^2+\lambda/\lambda_{\min}(\hat{\vc\Sigma})}{1+\lambda/\lambda_{\min}(\hat{\vc\Sigma})}
=1-\frac{1-2\vc\Xi^2}{1+\lambda/\lambda_{\min}(\hat{\vc\Sigma})}
\leq 1-\frac{1-\min\{1/2,C_2p/n\}}{1+C_3\lambda}.
\ee
From (\ref{eq4-11})-(\ref{eq4-12}), we have
\bes\label{eq4-13}
\left\|\hat{\vc\Sigma}^{1/2}(\vtheta_{t+1}-\hat{\vtheta})\right\|_2&\leq& \frac{2\vc\Xi^2+\lambda/\lambda_{\min}(\hat{\vc\Sigma})}{1+\lambda/\lambda_{\min}(\hat{\vc\Sigma})}
\left\|\hat{\vc\Sigma}^{1/2}(\vtheta_t-\hat{\vtheta})\right\|_2
+\mathcal{O}_p(N^{-1/2})+\mathcal{O}_p(\Delta_{nm\alpha})\nonumber\\
&\leq&\sbr{1-\frac{1-\min\{1/2,C_2p/n\}}{1+C_3\lambda}}\left\|\hat{\vc\Sigma}^{1/2}(\vtheta_t-\hat{\vtheta})\right\|_2\nonumber\\
&&+\mathcal{O}_p(N^{-1/2})+\mathcal{O}_p(\Delta_{nm\alpha})\nonumber\\
&\leq&\sbr{1-\frac{1-\min\{1/2,C_2p/n\}}{1+C_3\lambda}}^{t+1}\left\|\hat{\vc\Sigma}^{1/2}(\vtheta_0-\hat{\vtheta})\right\|_2\nonumber\\
&&+\mathcal{O}_p(N^{-1/2})+\mathcal{O}_p(\Delta_{nm\alpha}).
\ees
In addition, from (ii), we have $-\frac{1}{2}\vc\Sigma\preceq \hat{\vc\Sigma}-\vc\Sigma\preceq\frac{1}{2}\vc\Sigma$, and then $\frac{1}{2}\vc\Sigma\preceq \hat{\vc\Sigma}\preceq\frac{3}{2}\vc\Sigma$. Therefore,
$$\frac{1}{2}\lambda_{\min}(\vc\Sigma)\leq \lambda_{\min}(\hat{\vc\Sigma})\leq \lambda_{\max}(\hat{\vc\Sigma})\leq \frac{3}{2}\lambda_{\max}(\vc\Sigma).$$
Thus,
\be\label{eq4-14}
\frac{\lambda_{\max}(\hat{\vc\Sigma})}{\lambda_{\min}(\hat{\vc\Sigma})}
\leq\frac{3\lambda_{\max}(\vc\Sigma)}{\lambda_{\min}(\vc\Sigma)}=3\kappa.
\ee
From (\ref{eq4-13})-(\ref{eq4-14}), we have
\bes
\left\|(\vtheta_{t}-\hat{\vtheta})\right\|_2&\leq&
\sbr{1-\frac{1-\min\{1/2,C_2p/n\}}{1+C_3\lambda}}^t\sbr{\frac{\lambda_{\max}(\hat{\vc\Sigma})}{\lambda_{\min}(\hat{\vc\Sigma})}}^{1/2}\left\|(\vtheta_0-\hat{\vtheta})\right\|_2\nonumber\\
&&+\mathcal{O}_p(N^{-1/2})+\mathcal{O}_p(\Delta_{nm\alpha})\nonumber\\
&\leq&\sqrt{3\kappa}\sbr{1-\frac{1-\min\{1/2,C_2p/n\}}{1+C_3\lambda}}^t\left\|(\vtheta_0-\hat{\vtheta})\right\|_2\nonumber\\
&&+\mathcal{O}_p(N^{-1/2})+\mathcal{O}_p(\Delta_{nm\alpha}).\nonumber
\ees

We complete the proof (a).

\end{document}